%% file: main.tex
\documentclass[10pt,twocolumn,letterpaper]{article}

\usepackage[pagenumbers]{cvpr}

\input{preamble}

\definecolor{cvprblue}{rgb}{0.21,0.49,0.74}
\usepackage[pagebackref,breaklinks,colorlinks,allcolors=cvprblue]{hyperref}

\def\confName{CVPR}
\def\confYear{2027}

\title{Disentangling Policy Design in Task-Driven MRI Acquisition}

\author{
Lemuel Puglisi$^{1,2}$\thanks{Work conducted during an internship at Microsoft Research.}
\quad Olivier Jaubert$^{1}$\quad Brett Levac$^{1}$\quad Sebastiano Battiato$^{2}$\\[3pt]
Fernando P\'erez-Garc\'ia$^{1}$\quad Michael S. Hansen$^{1}$\quad Sam Bond-Taylor$^{1}$\\[6pt]
$^{1}$Microsoft Research, Cambridge, UK\qquad
$^{2}$University of Catania, Catania, Italy
}

\begin{document}

% This is to avoid the sections from the main paper
% appearing in the Appendix TOC.
\etocdepthtag.toc{paper}

\maketitle

%========================================
% ABSTRACT
%========================================

\begin{abstract}
Task-driven magnetic resonance imaging (MRI) aims to reduce scan time by prioritizing measurements relevant to automated diagnostic prediction. Recent methods use reinforcement learning to optimize acquisition policies that guide the sequential selection of measurement readouts, yet how policy design shapes the trade-off between acquisition cost and diagnostic performance remains unclear. In this paper, we examine three key design axes in a controlled setting: (i) policy inputs, comparing patient-dependent readout selection with learned sequences shared across scans; (ii) policy optimization, comparing greedy optimization of immediate rewards with longer-horizon strategies; and (iii) action space, evaluating different readout geometries, such as Cartesian lines and radial spokes, across diagnostic tasks. We also investigate unifying these geometries into a shared action space, termed a Bank of Readouts (BoR), where policies can learn which geometries to use and how to combine them for each task. We evaluate these choices on three pathology prediction tasks on the NYU fastMRI knee and brain datasets at fixed acceleration factors and under confidence-based stopping. Our results provide three main takeaways: (i) in contrast to reported benefits in prior work, patient-dependent inputs offer no consistent advantage over learned shared sequences; (ii) greedy learning performs competitively with longer-horizon optimization; and (iii) different tasks favor different readout geometries, while BoR is competitive with the strongest single-geometry alternatives. We believe that our findings help identify which aspects of policy design yield measurable gains and prioritize directions for future research.
\end{abstract}

%========================================
% INTRODUCTION
%========================================

\begin{figure}[t]
    \centering
    \includegraphics[width=0.9\linewidth]{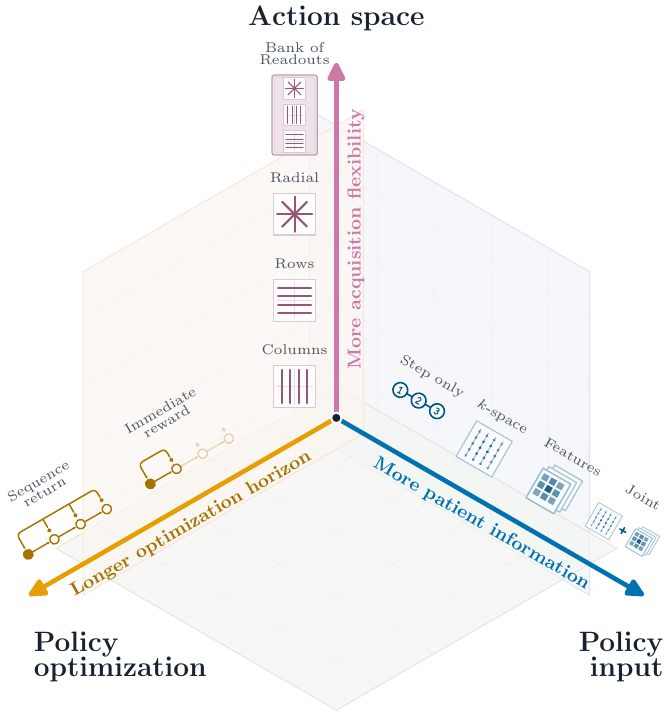}
    \caption{\textbf{Design space of task-driven MRI acquisition.} We examine three axes: what the policy observes, how it is optimized, and which readouts it can select.}
    \label{fig:teaser}
\end{figure}

\section{Introduction}
\label{sec:introduction}

Magnetic resonance imaging (MRI) provides detailed soft-tissue contrast for detecting anatomical and physiological abnormalities~\cite{chalela2007magnetic}, yet lengthy acquisitions remain a barrier to timely and accessible diagnosis~\cite{mclean2023mri}. A major contributor to scan time is the sequential sampling of \emph{$k$-space}, the spatial-frequency domain from which images are reconstructed. Scanners collect $k$-space measurements in groups termed \emph{readouts}, each comprising samples along a trajectory. Reducing the number of readouts is therefore a primary avenue for accelerating MRI. Although parallel imaging~\cite{deshmane2012parallel}, compressed sensing~\cite{lustig2007sparse}, and learned reconstruction~\cite{wang2016accelerating,jalal2021robust} can recover useful images from fewer readouts, they remain anchored to the conventional pipeline: data are acquired for image reconstruction, and diagnostic interpretation follows afterward. Recent work has instead explored \emph{task-driven acquisition}, selecting readouts for automated prediction of targets such as pathology~\cite{du2024mri,yen2024adaptive} or quantitative measurements~\cite{fischer2025cute,du2025active}. By prioritizing task-informative measurements early in the scan, this approach aims to collect sufficient diagnostic evidence using fewer readouts.

For pathology prediction, prior work formulates readout selection as a sequential decision problem and uses reinforcement learning (RL) to train acquisition policies that maximize a classification reward~\cite{yen2024adaptive,du2024mri}. However, differences in policy formulation and baseline comparisons make it difficult to isolate the effects of individual design choices. First, conditioning readout selection on previously acquired measurements~\cite{yen2024adaptive,du2024mri} enables patient-dependent adaptation. Yet these policies are rarely compared with non-adaptive counterparts trained under the same setup, so the gains from this adaptation remain unclear. Second, existing approaches maximize either immediate step-wise rewards~\cite{du2024mri} or cumulative future rewards~\cite{yen2024adaptive}. The benefit of long-horizon planning remains untested under controlled conditions. Finally, these frameworks restrict readout geometries to Cartesian lines~\cite{yen2024adaptive,du2024mri}, leaving non-Cartesian alternatives unexplored. Together, these gaps raise a central question: \emph{Which design choices matter for task-driven MRI acquisition?}

In this work, we address this question in a controlled study of acquisition policy design for automated pathology prediction. As illustrated in Figure~\ref{fig:teaser}, we examine three key design axes, varying one at a time within a unified retrospective MRI environment and evaluation framework:

\begin{itemize}

    \item \textbf{Policy inputs.} We assess the added value of patient-dependent readout selection. We compare policies that use acquired $k$-space measurements~\cite{yen2024adaptive}, classifier features~\cite{du2024mri}, or both with policies that receive no patient-dependent inputs and learn acquisition sequences shared across scans.
    
    \item \textbf{Policy optimization.} We examine whether accounting for future diagnostic rewards improves upon greedy optimization of immediate rewards. We compare greedy policy-gradient training~\cite{bakker2020experimental} with longer-horizon strategies based on proximal policy optimization (PPO)~\cite{schulman2017proximal} and Rainbow~\cite{hessel2018rainbow}.
    
    \item \textbf{Action space.} We investigate whether different pathology prediction tasks favor different readout geometries by comparing Cartesian columns, Cartesian rows, and radial spokes. We propose a \emph{Bank of Readouts} (BoR) that unifies these geometries into a shared action space, allowing policies to select and combine them for each task.

\end{itemize}

We evaluate these design choices on three pathology prediction tasks in knee and brain MRI. Beyond conventional comparisons at fixed acceleration factors, our evaluation framework incorporates confidence-based early stopping. This allows us to (i) trace acquisition-performance curves across stopping thresholds to relate diagnostic performance to the mean number of acquired readouts, and (ii) examine adaptivity due to confidence-based stopping separately from adaptivity due to patient-dependent readout selection.

%========================================
% RELATED WORKS
%========================================

\section{Related Work}
\label{sec:related-work}

\paragraph{Population-level sampling optimization.}
Sampling patterns can be optimized before acquisition and shared across subjects. LOUPE~\cite{bahadir2020deep,zhang2020extending} and DPS~\cite{huijben2020deep} learn Cartesian undersampling patterns jointly with reconstruction networks through probabilistic formulations. Non-Cartesian methods expand the available sampling geometries: SPARKLING~\cite{lazarus2019sparkling} designs variable-density trajectories under gradient and slew-rate constraints, while BJORK~\cite{wang2022b} jointly optimizes spline-based trajectories and reconstruction networks. SNOPY~\cite{wang2023stochastic} extends trajectory optimization to 3D acquisitions under physical constraints. While yielding optimized static acquisition patterns, these methods cannot exploit any potential gains that might arise from tailoring readout selection to individual scans.

\paragraph{Adaptive acquisition for reconstruction.}
Adaptive acquisition methods explore the potential benefits of patient-dependent sampling by selecting measurements using information collected during the scan. Methods targeting reconstruction quality include tree-search supervision~\cite{jin2019self}, active probabilistic subsampling~\cite{van2021active}, and RL for Cartesian~\cite{pineda2020active,bakker2020experimental} or radial acquisition~\cite{xu2026adaptive}. Beyond developing acquisition policies, several studies also examine how policy design affects reconstruction quality. Bakker et al.~\cite{bakker2020experimental} find that greedy learning remains competitive with longer-horizon optimization and that patient-dependent readout selection improves reconstruction. Evidence for the latter is mixed, however: shared acquisition sequences can outperform policies using patient-dependent inputs~\cite{pineda2020active}, and policies with access to acquired measurements can learn to ignore their inputs~\cite{bakker2022learning}. These findings motivate evaluating optimization and policy inputs as separate design choices.

\paragraph{Task-driven MRI acquisition.}
Task-driven methods select measurements based on their value for downstream prediction. Wu et al.~\cite{wu2024learning} jointly optimize a shared sampling pattern and a prediction network for segmentation and pathology classification. Other works extend task-driven acquisition to the patient-dependent setting: Du et al.~\cite{du2024mri} use greedy policy-gradient training on classifier features from partial reconstructions, whereas ASMR~\cite{yen2024adaptive} optimizes a policy via PPO on acquired $k$-space measurements, reporting gains over shared sequences built via static greedy search. Du et al.~\cite{du2025active} further extend adaptive sampling to sequential diagnostic tasks via multi-objective RL. As these RL methods differ in inputs, optimization horizons, and baselines, individual effects are hard to disentangle. Our work decouples these factors in a controlled setting and extends the action space beyond Cartesian geometries.

%========================================
% METHODOLOGY
%========================================

\section{Methodology}
\label{sec:methods}
We first define the retrospective MRI acquisition environment and the diagnostic classifier used for prediction and reward computation. We then describe the three policy-design axes: inputs, optimization, and action space, introducing BoR to support selection across readout geometries. Finally, we describe the confidence-based stopping strategy.

\subsection{Retrospective MRI Environment}
\label{sec:environment}

\begin{figure}[t]
    \centering
    \includegraphics[width=\linewidth]{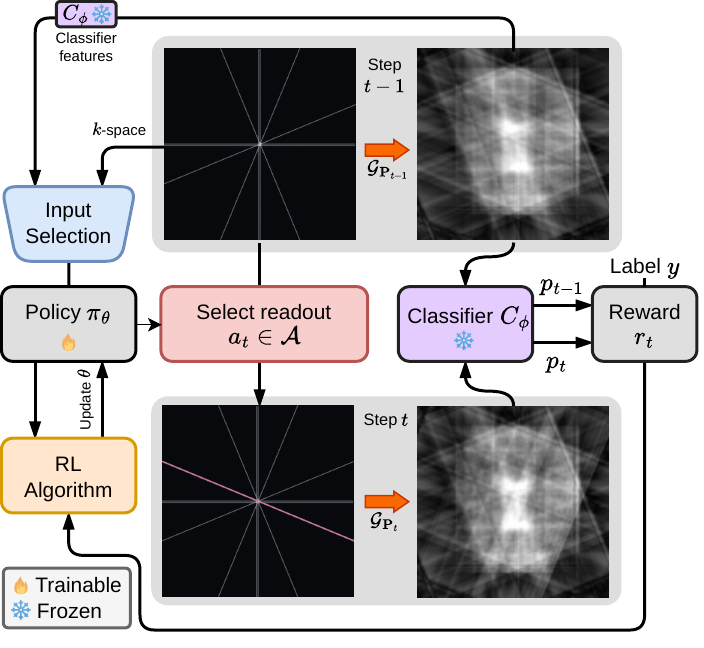}
    \caption{\textbf{MRI environment overview.} The policy selects a readout from the available action space using the configured inputs. Acquired measurements update the adjoint image, and the frozen diagnostic classifier produces predictions before and after acquisition. The reduction in classification loss defines reward $r_t$, which guides policy optimization.}
    \label{fig:method-pipeline}
\end{figure}

As illustrated in Figure~\ref{fig:method-pipeline}, our retrospective environment models MRI acquisition as a sequential decision loop. At each acquisition step, the policy uses its selected inputs to choose a readout, the newly acquired measurements are used to update the diagnostic prediction, and the training reward reflects the readout's effect on that prediction.

We use fully sampled MRI data and a single-coil Fourier model to retrospectively generate the acquired measurements. Let $\mathbf{x}\in\mathbb{C}^{H\times W}$ denote a complex MRI slice and $y\in\{0,1\}$ its pathology label. Each action $a\in\mathcal{A}$ selects a predefined readout comprising $L$ Fourier measurements at coordinates $\mathbf{P}(a)\in\mathbb{R}^{2\times L}$, with each column specifying one spatial-frequency location. The action space $\mathcal{A}$ determines the available readouts and their geometries (\cref{sec:action-space}). Acquisition starts from an empty collection of sampling coordinates, $\mathbf{P}_0=\varnothing$. At step $t$, selecting action $a_t$ appends the corresponding readout coordinates:
\begin{equation}
    \mathbf{P}_t
    =[\mathbf{P}_{t-1},\mathbf{P}(a_t)] \in\mathbb{R}^{2\times tL},
    \label{eq:trajectory}
\end{equation}
where brackets denote concatenation. We compute the corresponding measurements $\mathbf{q}_t\in\mathbb{C}^{tL}$ and adjoint image $\hat{\mathbf{x}}_t\in\mathbb{C}^{H\times W}$ as
\begin{equation}
    \mathbf{q}_t
    =\mathcal{F}_{\mathbf{P}_t}\mathbf{x},
    \qquad
    \hat{\mathbf{x}}_t
    =\mathcal{G}_{\mathbf{P}_t}(\mathbf{q}_t),
    \label{eq:acquisition}
\end{equation}
where $\mathcal{F}_{\mathbf{P}_t}$ evaluates the Fourier transform at the acquired coordinates and $\mathcal{G}_{\mathbf{P}_t}$ maps the measurements to the image grid using an adjoint-based reconstruction. We summarize the acquisition history as $h_t=(\mathbf{q}_t,\mathbf{P}_t,\hat{\mathbf{x}}_t)$. A policy $\pi_{\boldsymbol{\theta}}$, with learnable parameters $\boldsymbol{\theta}$, selects the next readout according to
$a_t\sim\pi_{\boldsymbol{\theta}}(\cdot\mid h_{t-1})$. The policy's input configuration determines its observations (\cref{sec:selection}). We mask previously selected actions to prevent reselection~\cite{pineda2020active} and acquire at most $T$ readouts, where $T$ denotes the acquisition budget. During training, each action receives a reward $r_t$ based on the reduction in classification loss of a frozen diagnostic classifier (\cref{sec:classifier}). We detail the implementation of the operators in Eq.~\eqref{eq:acquisition} in Appendix~\appref{app:acquisition}{A}.

\subsection{Diagnostic Classifier}
\label{sec:classifier}
Following~\cite{yen2024adaptive,du2024mri}, we train a diagnostic classifier $C_{\boldsymbol{\phi}}$ to predict pathology from partial acquisitions and use its predictions to compute rewards for policy training. Given a training slice $\mathbf{x}$ with label $y$, we generate a partial acquisition by randomly sampling a readout budget $T$ and a corresponding collection of sampling coordinates $\mathbf{P}_T$. We then compute the adjoint image $\hat{\mathbf{x}}=\mathcal{G}_{\mathbf{P}_T}(\mathcal{F}_{\mathbf{P}_T}\mathbf{x})$ and train the classifier to predict the pathology probability $C_{\boldsymbol{\phi}}(\hat{\mathbf{x}})\in[0,1]$ by minimizing the expected binary focal loss~\cite{lin2017focal}:
\begin{equation}
    \mathbb{E}_{(\mathbf{x},y),\,T,\,\mathbf{P}_T}
    \left[
        \mathcal{L}_{\mathrm{focal}}
        \bigl(C_{\boldsymbol{\phi}}(\hat{\mathbf{x}}),y\bigr)
    \right].
    \label{eq:classifier-training}
\end{equation}
To expose the classifier to different readout geometries and their combinations, we sample $\mathbf{P}_T$ from individual readout families and their mixtures. Specifically, with probability $\rho_{\mathrm{mix}}$, we sample family proportions from a Dirichlet distribution and allocate the budget $T$ across Cartesian columns, Cartesian rows, and radial spokes. Otherwise, we randomly select one family and sample all $T$ readouts from it. We provide further sampling and training settings in Appendix~\appref{app:classifier-training}{C.1}.

\subsection{Policy Architecture}
\label{sec:policy-architecture}
We parameterize the acquisition policy with a transformer backbone~\cite{vaswani2017attention}. We employ modality-specific tokenizers to encode acquired $k$-space measurements and spatial classifier features into two token sequences. The transformer integrates these sequences through joint attention~\cite{esser2024scaling}, while an acquisition-step embedding conditions each block through adaptive layer normalization~\cite{peebles2023scalable}. When only one modality is available, attention operates within its token sequence. An output head maps the resulting representation to scores over the available readouts. We detail the tokenizers and network architecture in Appendices~\appref{app:tokenizers}{C.2} and \appref{app:policy-architecture}{C.3}, respectively.

\subsection{Policy Inputs}
\label{sec:selection}
To analyze the policy-input axis, we vary the observations supplied through the \emph{Input Selection} block in Figure~\ref{fig:method-pipeline}, selectively disabling the $k$-space and classifier-feature token streams while retaining the acquisition-step embedding. We train a separate policy for each of four configurations:
\begin{itemize}
    \item $\bm{k}$\textbf{-space only}: inspired by~\cite{yen2024adaptive}, the policy uses the acquired complex measurements $\mathbf{q}_t$ and their sampling coordinates $\mathbf{P}_t$.
    \item \textbf{Features only}: inspired by~\cite{du2024mri}, the policy uses spatial features extracted by the frozen classifier $C_{\boldsymbol{\phi}}$ from the current adjoint image $\hat{\mathbf{x}}_t$.
    \item \textbf{Joint inputs}: the policy uses both modalities.
    \item \textbf{Step only}: the policy uses only the acquisition-step embedding.
\end{itemize}
Without patient-dependent inputs, the step-only policy learns a task-specific acquisition sequence shared across slices.

\subsection{Policy Optimization}
\label{sec:optimization-method}
To analyze the optimization axis, we vary the \emph{RL Algorithm} block in Figure~\ref{fig:method-pipeline}. All methods share the same reward, defined as the reduction in binary cross-entropy loss after acquiring a readout:
\begin{equation}
    r_t
    =\mathcal{L}_{\mathrm{BCE}}(p_{t-1},y)
     -\mathcal{L}_{\mathrm{BCE}}(p_t,y),
    \label{eq:reward}
\end{equation}
where $p_t=C_{\boldsymbol{\phi}}(\hat{\mathbf{x}}_t)$ is the pathology probability predicted by the frozen classifier after $t$ readouts (with $p_0 = C_{\boldsymbol{\phi}}(\mathbf{0})$). We compare multi-sample greedy policy-gradient training~\cite{bakker2020experimental,du2024mri}, which maximizes immediate reward, with two algorithms that optimize for longer-horizon returns: PPO~\cite{schulman2017proximal} and Rainbow~\cite{hessel2018rainbow}. Training details for all three algorithms are provided in Appendix~\appref{app:optimization}{D}.

\subsection{Action Space}
\label{sec:action-space}
\begin{figure}[t]
    \centering
    \includegraphics[width=\linewidth]{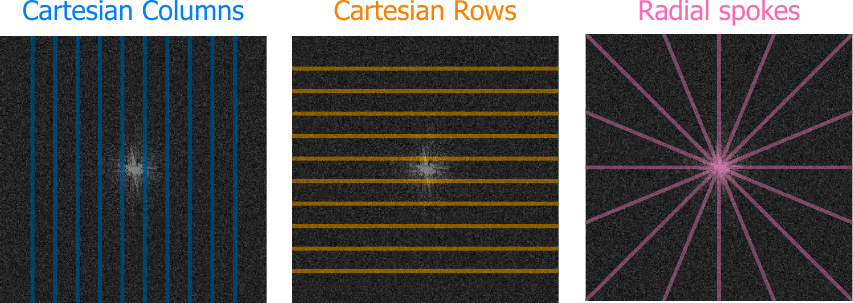}
    \caption{\textbf{Readout geometries.} We illustrate subsets of Cartesian columns, Cartesian rows, and radial spokes in $k$-space.}
    \label{fig:readout-geometries}
\end{figure}

To analyze the action-space axis, we investigate whether different pathologies favor different readout geometries and whether combining geometries provides additional value. We introduce BoR, which combines Cartesian columns, Cartesian rows, and radial spokes (illustrated in Figure~\ref{fig:readout-geometries}) within a single discrete action space. Each of the $M=3$ geometries contributes $J$ candidate readouts. An action $a=(m,j)$ selects readout $j$ from geometry $m$, defining
\begin{equation}
    \mathcal{A}_{\mathrm{BoR}}=[M]\times[J],
    \qquad
    |\mathcal{A}_{\mathrm{BoR}}|=MJ,
    \label{eq:bor}
\end{equation}
where $[n]=\{1,\ldots,n\}$. We compare four action-space configurations in the \emph{Select readout} block of Figure~\ref{fig:method-pipeline}: \emph{Columns}, \emph{Rows}, and \emph{Radial}, which restrict selection to the corresponding geometry, and \emph{BoR}, which permits selection across all three geometries. We restrict our study to Cartesian lines and radial spokes, for which prior work reports similar acquisition times at matched readout counts~\cite{van2023cartesian}. Since each readout contains $L$ measurements, equal readout counts ensure equal measurement budgets across geometries. We detail the readout parameterizations in Appendix~\appref{app:acquisition}{A}.

\subsection{Confidence-Based Stopping}
\label{sec:stopping}
Following prior work on early stopping in MRI~\cite{wen2024task,fischer2025cute}, we allow the diagnostic classifier to trigger early termination at inference time when its confidence reaches a chosen threshold. At step $t$, let $\tilde{p}_t$ denote a calibrated version of the classifier output $p_t$. We define classification confidence as
\begin{equation}
    c_t
    =\max\{\tilde{p}_t,1-\tilde{p}_t\}.
    \label{eq:confidence}
\end{equation}
Let $t_{\min}$ be the earliest acquisition step at which stopping is allowed. For a confidence threshold
$\eta\in[0.5,1]$, we define the stopping time as
\begin{equation}
    \tau(\eta)
    =\min\!\left(
    \{t\in\{t_{\min},\ldots,T\}:c_t\geq\eta\}\cup\{T\}
    \right).
    \label{eq:stopping}
\end{equation}
Starting at $t_{\min}$, we stop acquisition at the first step with
$c_t\geq\eta$, or at the maximum budget $T$ if the threshold is
never met. We use $\tilde{p}_{\tau(\eta)}$ as the final prediction.

We fit the per-step calibration maps from $p_t$ to $\tilde{p}_t$ on the validation set so that predicted probabilities better match observed outcome frequencies. We provide calibration details in Appendix~\appref{app:calibration}{E}.

%========================================
% EXPERIMENTS
%========================================

\section{Experiments}
\label{sec:experiments}
We first describe the datasets, experimental setup, and evaluation protocol. We then conduct controlled comparisons across the three design axes to evaluate their impact on pathology prediction, and examine whether our findings on policy inputs extend to the original ASMR~\cite{yen2024adaptive} setting. Next, we analyze how the number of readouts acquired under confidence-based stopping varies with pathology status and slice position. Finally, we benchmark our default configuration against existing acquisition baselines.

\subsection{Data and Tasks}
\label{sec:data}
We evaluate three binary pathology-detection tasks using 2D slices from the NYU fastMRI Initiative~\cite{knoll2020fastmri,zbontar2018fastmri} with fastMRI+ annotations~\cite{zhao2022fastmri+}: meniscus tear, joint effusion, and brain abnormalities. All splits are patient-disjoint to prevent leakage. The knee cohort contains \num{41877} slices, with \num{30562}, \num{4180}, and \num{7135} slices in the training, validation, and test sets, respectively. Both knee tasks use the same splits. The brain cohort contains \num{7513} slices, with \num{4677}, \num{975}, and \num{1861} slices in the training, validation, and test sets, respectively. For the brain task, we follow the label grouping and slice-filtering procedure of Schlimbach et al.~\cite{schlimbach2026generative}, grouping seven abnormalities into one positive class. Appendix~\appref{app:datasets}{B.1} provides label-assignment rules and further dataset details.

\subsection{Experimental Setup}
\label{sec:experimental-setup}
We use common training and evaluation settings across policy configurations unless otherwise specified. We resize all 2D MRI slices to $H=W=128$ and use $L=128$ equally spaced $k$-space measurements per readout across all geometries. Each acquisition starts from an empty history, and we define nominal acceleration relative to a reference count of $T_{\max}=128$ readouts. We repeat each experiment with five random seeds. For each task and seed, we train a ResNet-50 diagnostic classifier~\cite{he2016deep} and reuse its frozen checkpoint across all policy-design configurations. We match classifier and policy training seeds and select checkpoints using validation data only. We detail further implementation and training settings in Appendices~\ref{app:networks} and~\ref{app:optimization}.

\paragraph{Default configuration.} We use joint inputs and BoR to give the policy access to both input modalities and all readout geometries, and adopt greedy policy-gradient training following prior work~\cite{bakker2020experimental,du2024mri}. Each policy-design experiment varies one axis while holding the other two fixed.

\subsection{Evaluation Protocol}
\label{sec:evaluation}
We use slice-level area under the precision-recall curve (AUPRC) as the primary metric because all tasks exhibit class imbalance. We report additional metrics and tabulated results in Appendix~\appref{app:add-results}{H}. For each metric, we report the mean and standard deviation (SD) across the five seeds. We consider two evaluation settings: fixed measurement budgets and confidence-based stopping.

\paragraph{Fixed measurement budgets.}
Following~\cite{du2024mri,yen2024adaptive}, we evaluate nominal acceleration factors $R\in\{4,8,16,32\}$, with readout budgets $T=T_{\max}/R$. At each budget, we acquire exactly $T$ readouts per test slice and compute AUPRC across the test set using the resulting classifier probabilities.

\paragraph{Confidence-based stopping.}
\label{sec:stopping-evaluation}
We apply the stopping rule in \cref{sec:stopping} and vary the confidence threshold $\eta$ from \num{0.70} to \num{0.975} to evaluate the trade-off between diagnostic performance and acquisition cost. We allow stopping after at least $t_{\min}=2$ readouts and acquire at most \num{32}, corresponding to $R=4$. Let $\tau_i(\eta)$ denote the number of readouts acquired for test slice $i$. We quantify acquisition cost using Measurements-to-Confidence (MTC), the mean readout count across the $N_{\mathrm{test}}$ test slices:
\begin{equation}
    \operatorname{MTC}(\eta)
    =\frac{1}{N_{\mathrm{test}}}
     \sum_{i=1}^{N_{\mathrm{test}}}\tau_i(\eta).
    \label{eq:mtc}
\end{equation}
At each threshold, we compute AUPRC from the calibrated predictions $\tilde{p}_{i,\tau_i(\eta)}$ at the slice-specific stopping steps. We introduce acquisition-performance curves by sweeping $\eta$ and plotting $(\operatorname{MTC}(\eta),\operatorname{AUPRC}(\eta))$. MTC measures readout count rather than physical scan time.

\begin{figure}[t]
    \centering
    \includegraphics[width=\linewidth]{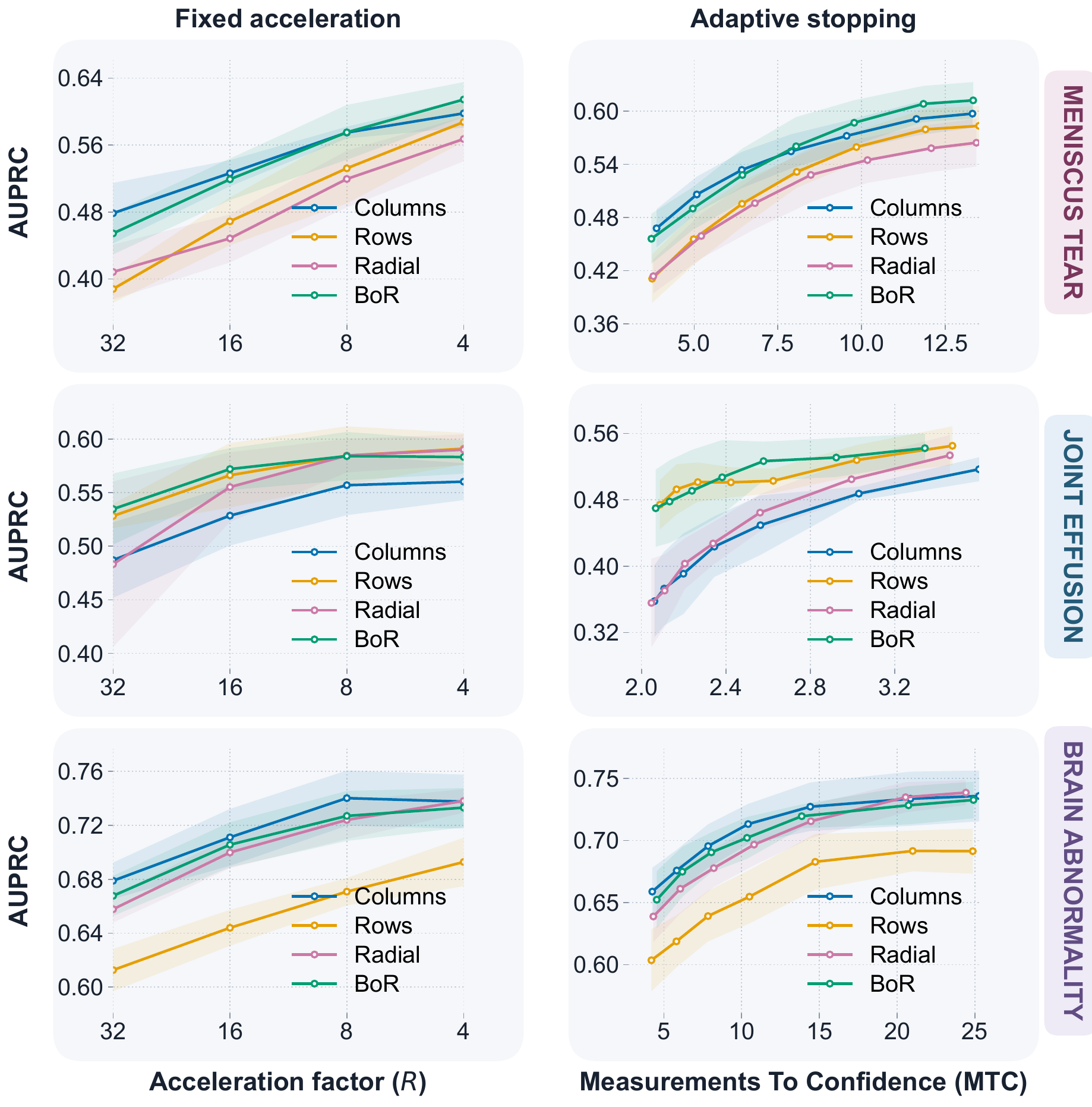}
    \caption{\textbf{Readout geometry.} AUPRC (mean $\pm$ SD) for individual readout families and BoR at fixed acceleration factors (left) and under confidence-based stopping (right).}
    \label{fig:readout-geometry}
\end{figure}

\subsection{Readout Geometry Is Task Dependent}
\label{sec:geometry-results}
We compare Columns, Rows, Radial, and BoR to investigate whether different pathologies favor different readout geometries and whether combining geometries provides additional value. Furthermore, we analyze the proportion of BoR readout selections assigned to each geometry at each acquisition step.

\begin{figure}[t]
    \centering
    \includegraphics[width=\linewidth]{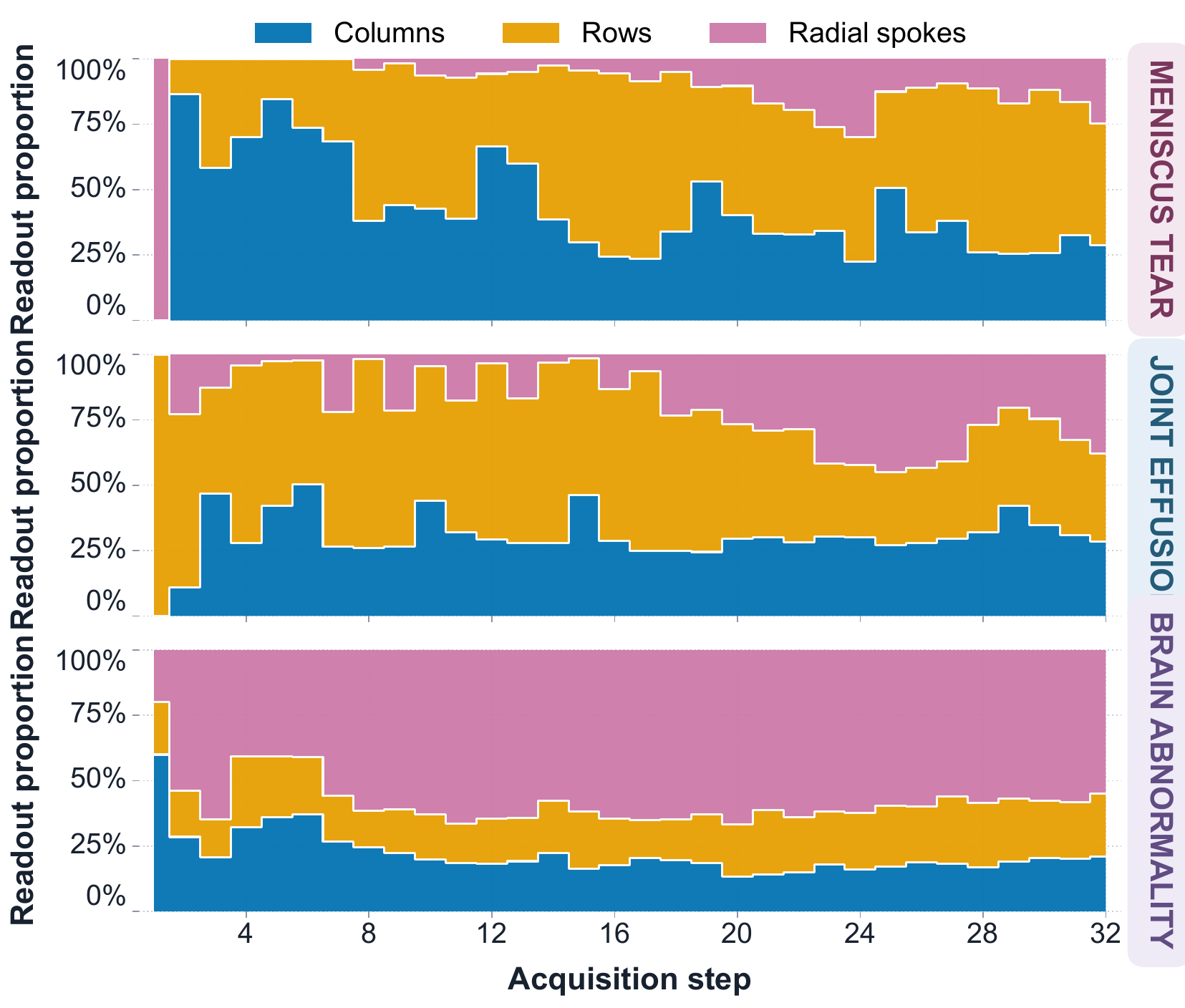}
    \caption{\textbf{Readout-family usage.} Proportion of readout selections assigned to Cartesian columns, Cartesian rows, and radial spokes at each acquisition step for the three pathology-prediction tasks.}
    \label{fig:family-usage}
\end{figure}

\paragraph{Results.}
Figure~\ref{fig:readout-geometry} shows that the strongest readout geometry depends on the diagnostic task: Columns generally performs best for meniscus-tear detection, Rows for joint-effusion detection, and Columns or Radial for brain-abnormality detection, depending on the budget. At $R=32$, Columns exceeds Radial by \num{7.0} percentage points (p.p.) in AUPRC for meniscus-tear detection, whereas Rows exceeds Columns by \num{4.1} p.p. for joint-effusion detection. Similar preferences appear under confidence-based stopping. BoR performs comparably to the strongest individual geometry across tasks. Figure~\ref{fig:family-usage} shows that its early readout selections generally align with these task-dependent preferences: meniscus-tear policies favor columns, joint-effusion policies favor rows, and brain-abnormality policies favor radial spokes. 

\subsection{Greedy Policy Learning Remains Competitive}
\label{sec:optimization-results}
We examine whether accounting for future diagnostic rewards improves readout selection compared with optimizing immediate rewards. We compare multi-sample greedy policy-gradient training~\cite{bakker2020experimental} with longer-horizon strategies based on PPO~\cite{schulman2017proximal} and Rainbow~\cite{hessel2018rainbow}.

\begin{figure}[t]
    \centering
    \includegraphics[width=\linewidth]{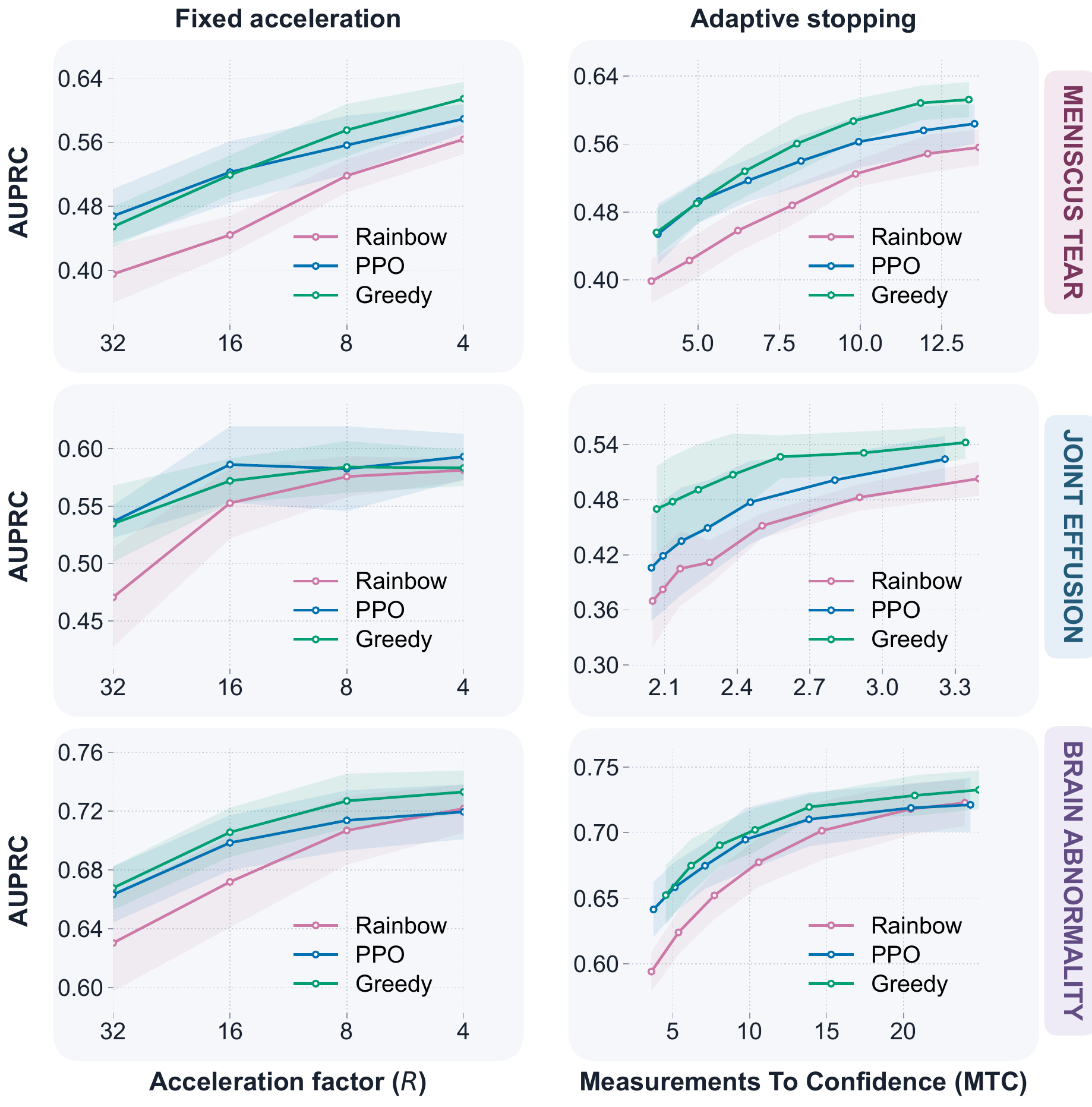}
    \caption{\textbf{Policy optimization.} AUPRC (mean $\pm$ SD) for greedy training, PPO, and Rainbow at fixed acceleration factors (left) and under confidence-based stopping (right).}
    \label{fig:rl}
\end{figure}

\paragraph{Results.}
Figure~\ref{fig:rl} shows that greedy training outperforms or remains competitive with PPO. At fixed budgets, PPO achieves higher mean AUPRC in some settings but provides no consistent improvement across tasks and accelerations. Under confidence-based stopping, greedy training generally matches or exceeds PPO at comparable MTC. Rainbow performs generally worse than both methods. 

\begin{figure}[t]
    \centering
    \includegraphics[width=\linewidth]{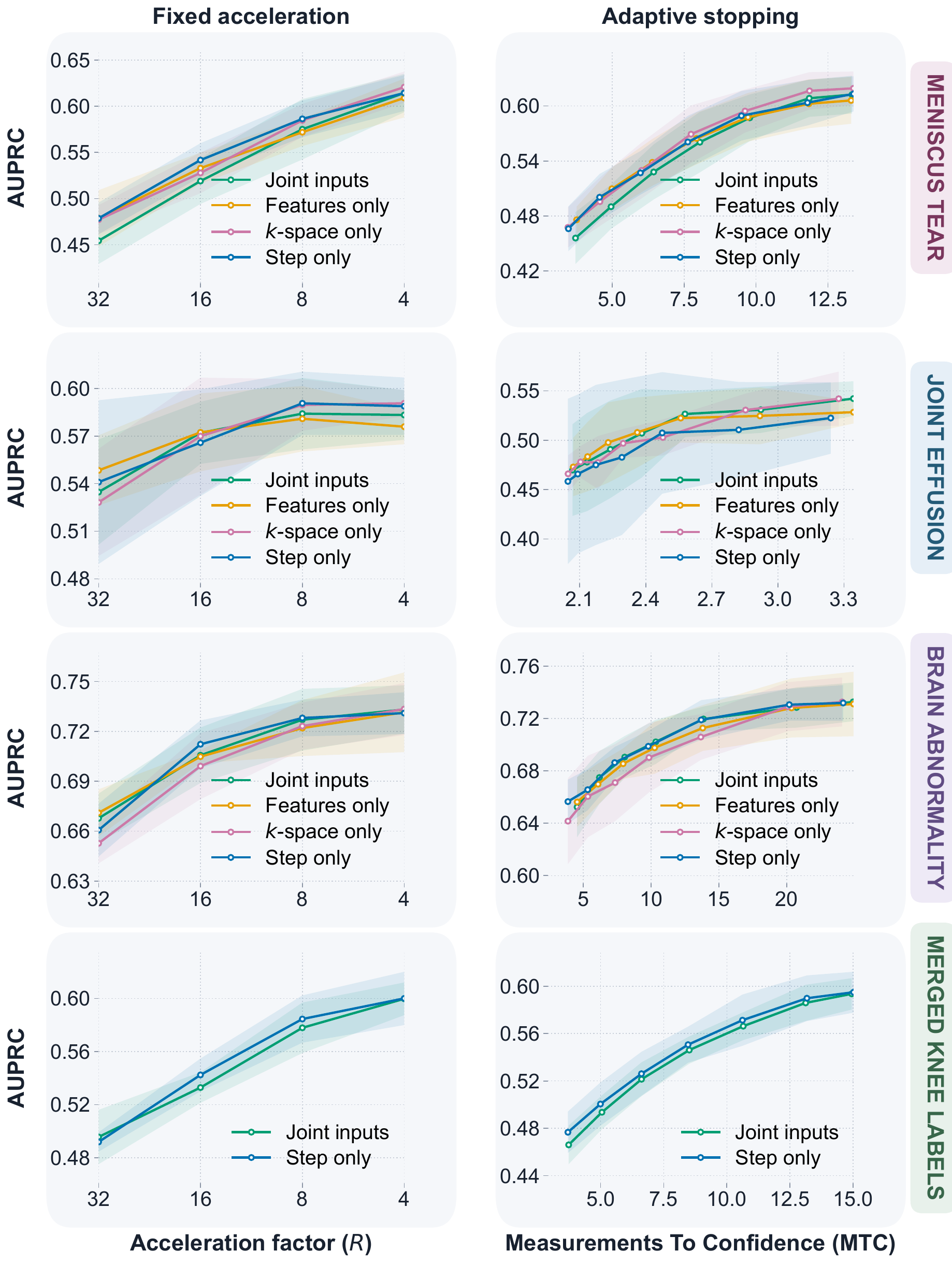}
    \caption{\textbf{Policy inputs.} AUPRC (mean $\pm$ SD) for four input configurations on the individual tasks (top three rows) and for joint versus step-only inputs on the combined knee task (bottom row). We evaluate fixed acceleration factors (left) and confidence-based stopping (right).}
    \label{fig:conditioning}
\end{figure}

\subsection{Shared Sequences Retain Strong Performance}
\label{sec:conditioning-results}
We examine whether patient-dependent information improves readout selection over learned shared sequences. First, we compare four policy-input configurations: $k$-space only, features only, joint inputs, and step only, where the last configuration learns a sequence shared across slices. Second, we test whether patient-dependent inputs help the policy select informative readouts based on early evidence of abnormalities that favor different readout geometries. To this end, we combine meniscus tear and joint effusion into a binary task, assigning a positive label when either abnormality is present, and compare joint- and step-only policies.

\paragraph{Results.}
Figure~\ref{fig:conditioning} shows no consistent gain from patient-dependent inputs over the step-only baseline at fixed measurement budgets: the step-only policy achieves higher mean AUPRC at several acceleration factors, while joint inputs offer no consistent improvement over either single modality. Under confidence-based stopping, all input configurations perform comparably on meniscus-tear and brain-abnormality detection. On joint-effusion detection, patient-dependent policies achieve higher mean AUPRC over part of the MTC range, whereas the step-only policy shows higher variance across seeds. On the combined knee task, the step-only policy matches the joint-input policy across both evaluation settings.

\subsection{Re-examining Patient Dependence in ASMR}
\label{sec:asmr-conditioning-results}
To assess whether the absence of a consistent benefit from patient-dependent inputs extends beyond our controlled setup, we ablate patient-dependent conditioning in ASMR~\cite{yen2024adaptive}, which reported gains over shared acquisition sequences constructed through static greedy search. Using ASMR's official implementation and original $640 \times 400$ complex $k$-space inputs, we compare the patient-dependent ASMR policy against a variant that receives only the remaining acquisition budget and learns a shared acquisition sequence. We train both variants with the original PPO procedure and the same frozen classifier within each seed, and evaluate them on ASMR's three knee targets (ACL, meniscus tear, and ``other abnormalities'') across five seeds. Alongside AUPRC, we report the area under the receiver operating characteristic curve (AUROC), ASMR's primary evaluation metric, and detail additional experimental settings in Appendix~\appref{app:asmr-ablation}{G}.

\begin{figure}[t]
    \centering
    \includegraphics[width=\linewidth]{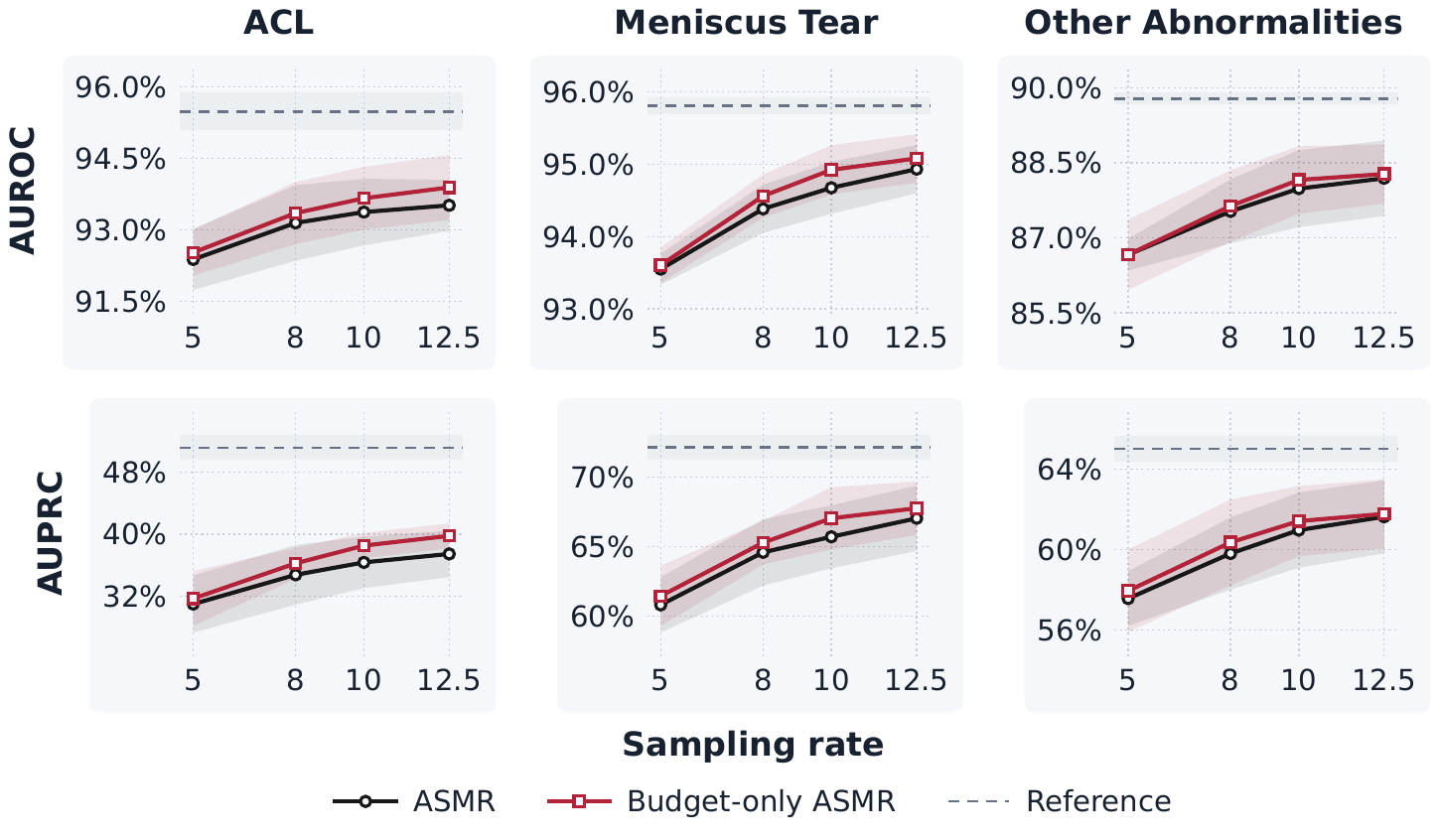}
    \caption{\textbf{Policy inputs in ASMR.} AUROC (top) and AUPRC (bottom), reported as mean $\pm$ SD, for the original ASMR policy and its budget-only variant on three knee targets at ASMR's sampling rates (\%). Dashed lines indicate the fully sampled classifier reference.}
    \label{fig:asmr-results}
\end{figure}

\paragraph{Results.}
Figure~\ref{fig:asmr-results} shows that the budget-only variant matches or exceeds the original policy in mean AUROC and AUPRC across all three targets. The differences are more pronounced for ACL and meniscus-tear detection, particularly in AUPRC, while both variants perform similarly on ``other abnormalities''. Together, these findings indicate no consistent advantage from patient-dependent readout selection over a learned shared sequence in the ASMR setting.

\begin{figure}[t]
    \centering
    \includegraphics[width=\linewidth]{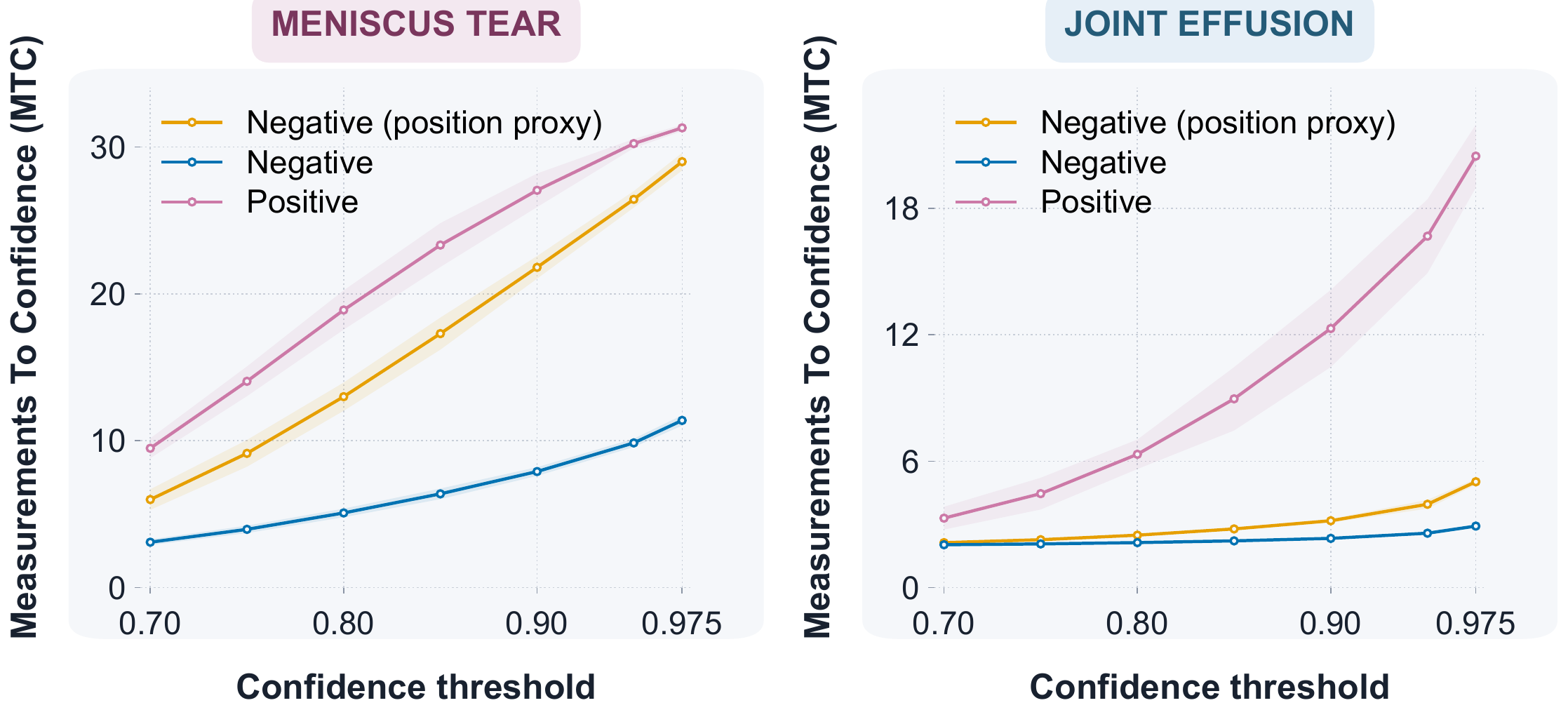}
    \caption{\textbf{Class-conditional stopping costs.} MTC (mean $\pm$ SD) across confidence thresholds for meniscus-tear and joint-effusion detection. We compare positive slices, all negative slices, and negative subsets within a position window derived from positive training slices.}
    \label{fig:class-conditional-mtc}
\end{figure}

\subsection{What Drives Confidence-Based Stopping?}
\label{sec:stopping-experiment}
We examine how confidence-based stopping allocates readouts across slices, focusing on differences associated with pathology status and slice position. For each knee task, we compare MTC across confidence thresholds for positive slices, all negative slices, and a subset of negative test slices within the 5th-95th percentile range of relative slice positions among positive training slices. This window provides a positional proxy for the region where positives typically occur, allowing us to assess how stopping costs relate to slice location. We restrict this analysis to the knee tasks because brain-abnormality detection combines findings with different anatomical distributions, making this positional proxy less well defined.

\paragraph{Results.}
Figure~\ref{fig:class-conditional-mtc} shows that positive slices receive more readouts than either negative group across both tasks. For meniscus-tear detection at $\eta=0.95$, positive slices receive an average of \num{30.22} readouts, compared with \num{9.85} for all negatives and \num{26.43} for negatives within the training-derived position window. Restricting negatives to this window increases their mean stopping cost, bringing it closer to that of positives. The difference between the two negative groups is less pronounced for joint-effusion detection, where positives still receive substantially more readouts than either group.

\begin{figure}[t]
    \centering
    \includegraphics[width=\linewidth]{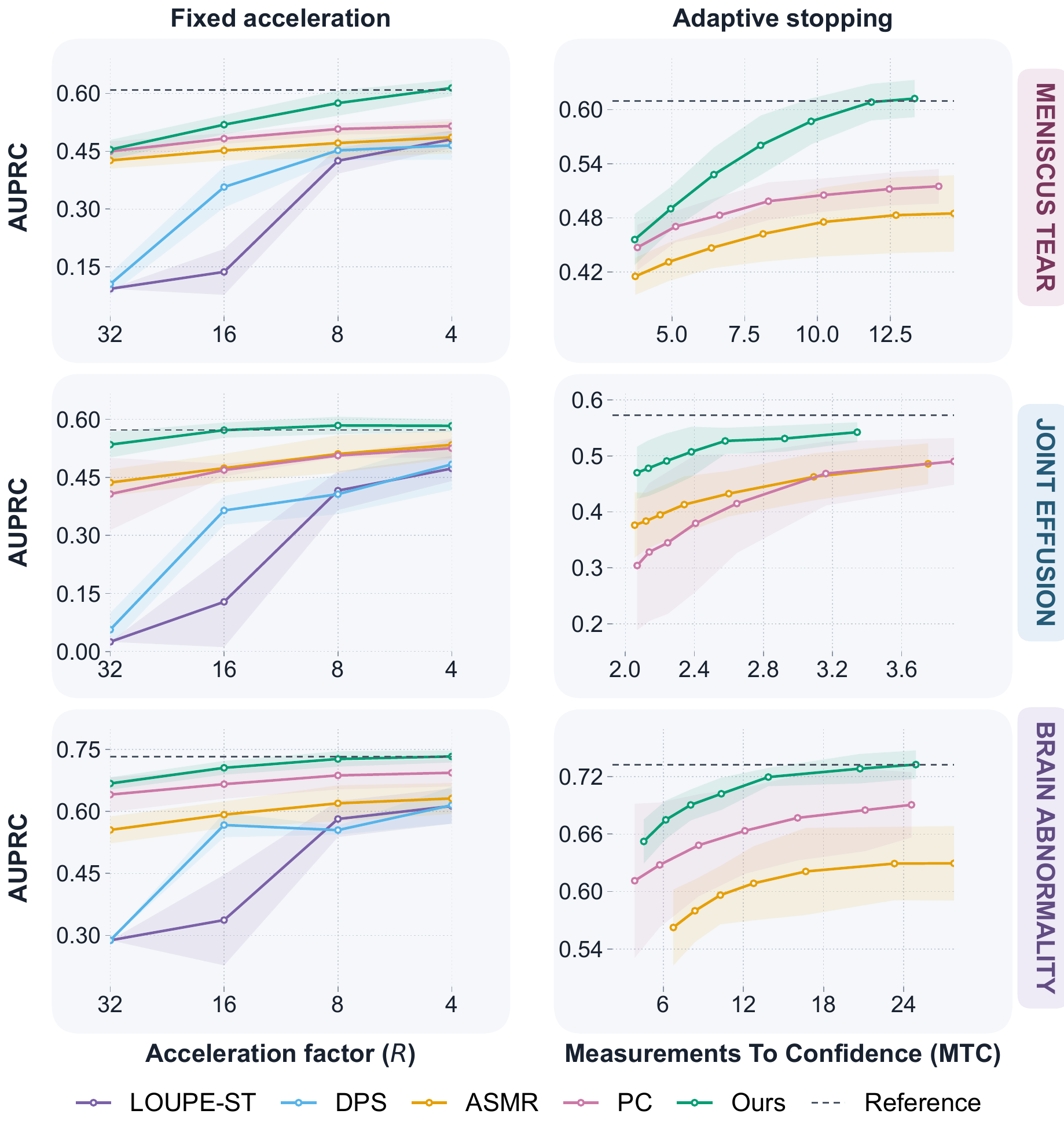}
    \caption{\textbf{Existing acquisition methods.} AUPRC (mean $\pm$ SD) of BoR and existing methods at fixed acceleration factors (left) and under confidence-based stopping (right). Fully sampled pathology prediction provides a reference.}
    \label{fig:sota}
\end{figure}

\subsection{Comparison with Existing Methods}
\label{sec:sota-results}
To complement our controlled experiments, we compare our default configuration against methods that learn shared sampling patterns---LOUPE-ST~\cite{zhang2020extending} (the straight-through variant of LOUPE~\cite{bahadir2020deep}) and DPS~\cite{huijben2020deep}---as well as adaptive readout-selection methods, namely ASMR~\cite{yen2024adaptive} and Policy Classifier (PC)~\cite{du2024mri}. We also include pathology prediction from fully sampled acquisitions as a reference, and detail all baseline implementations in Appendix~\appref{app:baselines}{F}.

\paragraph{Results.}
Figure~\ref{fig:sota} shows that our default configuration achieves the highest mean AUPRC among the evaluated methods at all four acceleration factors across the three tasks. Under confidence-based stopping, it also achieves higher mean AUPRC than ASMR and PC at comparable MTC over the overlapping portions of their curves. At higher confidence thresholds, mean AUPRC approaches or reaches the fully sampled reference for meniscus-tear and brain-abnormality detection.

%========================================
% DISCUSSION
%========================================

\section{Discussion}
\label{sec:discussion}
We conducted controlled comparisons to assess how readout geometry, policy optimization, and policy inputs affect pathology prediction in task-driven MRI acquisition. Our results suggest three considerations for designing and evaluating acquisition policies:

\begin{itemize}
    \setlength{\itemsep}{0.45em}
    \setlength{\parsep}{0pt}

    \item \emph{Different pathologies favor different readout geometries}, making the action space a key design choice. When multiple readout families are combined in BoR, we observe that the policy selects the favored geometry during the early acquisition phase. This flexibility allows the policy to be competitive with the best single-geometry policy without requiring prior geometry selection (\cref{sec:geometry-results}).

    \item \emph{Greedy optimization remains competitive with longer-horizon strategies} (\cref{sec:optimization-results}). Together with similar findings in MRI reconstruction~\cite{bakker2020experimental}, this suggests that optimizing immediate diagnostic rewards is a practical starting point for sequential readout selection.

    \item \emph{Patient-dependent readout selection offers no consistent advantage over learned shared sequences.} This finding also holds when the diagnostic target combines abnormalities that favor different readout geometries (\cref{sec:conditioning-results}). In ASMR's default settings, where patient-dependent selection was reported to outperform a shared sequence built by greedy search~\cite{yen2024adaptive}, we find that a variant receiving only the remaining acquisition budget matches or exceeds the original adaptive policy (\cref{sec:asmr-conditioning-results}). These findings highlight the need for strong learned shared-sequence baselines to assess whether the gains from patient-dependent selection justify its added complexity.

\end{itemize}

Finally, even when patient-dependent readout selection offers no consistent advantage, learned shared sequences still enable scan-specific acquisition lengths via confidence-based stopping, with readout counts varying systematically with pathology presence and slice position (\cref{sec:stopping-experiment}).

\paragraph{Limitations and Future Directions.}
While our study isolates key design choices in task-driven MRI acquisition, it also has limitations that point to directions for future research. First, following standard settings in task-driven acquisition~\cite{du2024mri,yen2024adaptive}, we focus on 2D slice-level classification under a single-coil acquisition model. Extending this analysis to multi-coil sensitivity models, volumetric 3D acquisitions, and other downstream tasks is an important next step. Second, following prior adaptive acquisition studies~\cite{bakker2020experimental,du2024mri,pineda2020active,yen2024adaptive}, our retrospective environment treats readout selections as independent of scanner transition constraints. Prospective deployment will require accounting for hardware and physical effects, such as gradient slew-rate limits and eddy currents, during readout selection.

\section*{LLM Usage Statement}
We used GitHub Copilot to assist with developing the research code and experiments, ChatGPT to refine the manuscript's writing and obtain feedback, and Gemini and Claude to obtain additional feedback. The authors reviewed and edited all assisted and generated content and take full responsibility for the entire submission.

{
    \small
    \bibliographystyle{ieeenat_fullname}
    \bibliography{main}
}

\clearpage
\input{appendix}

\end{document}

%% file: preamble.tex
\renewcommand{\paragraph}[1]{\vspace{.5em}\noindent\textbf{#1}}

\usepackage{bm}
\usepackage{siunitx}
\usepackage{booktabs}
\usepackage[table]{xcolor}

\usepackage{etoc}
\usepackage{minitoc}
\makeatletter
\newcommand{\appref}[2]{%
  \@ifundefined{r@#1}{#2}{\ref{#1}}%
}
\makeatother

%% file: appendix.tex
\appendix
\setcounter{tocdepth}{2}

\etocdepthtag.toc{supplement}
\setcounter{tocdepth}{2}
\etocsettagdepth{paper}{none}
\etocsettagdepth{supplement}{subsection}
\tableofcontents

\bigskip

%====================================
% Acquisition Environment
%====================================

\section{Acquisition Model Implementation}
\label{app:acquisition}

\subsection{Readout Coordinates}
\label{app:readout-coordinates}
We discretize each readout geometry into a finite set of candidate readouts, each corresponding to one action. We represent the $j$-th readout of family $m$ by its sampling coordinates $\mathbf{P}_j^{(m)}\in\mathbb{R}^{2\times L}$, so that $\mathbf{P}(a)=\mathbf{P}_j^{(m)}$ for $a=(m,j)$. Each column contains a pair $(k_x,k_y)^\top$, where $k_x$ and $k_y$ denote the horizontal and vertical spatial-frequency coordinates, respectively, expressed in radians per pixel. We use an image grid of size $H=W=128$ and $L=128$ measurements per readout. Each of the $M=3$ geometries contributes $J=128$ candidate readouts, yielding $MJ=384$ actions in BoR.

\paragraph{Cartesian columns and rows.}
We define Cartesian readouts using the centered Fourier bins
\begin{equation}
    b_j=\frac{2\pi}{J}(j-1-J/2),\qquad j=1,\ldots,J.
\end{equation}
For column $j$, we fix $k_x=b_j$ and sample $k_y$ at all $J$ bins. For row $j$, we fix $k_y=b_j$ and sample $k_x$ at all $J$ bins. Since $L=J=128$, the corresponding coordinate matrices are
\begin{equation}
    \begin{aligned}
        \mathbf{P}_j^{(\mathrm{col})}
        &=
        \begin{pmatrix}
            b_j & \cdots & b_j\\
            b_1 & \cdots & b_J
        \end{pmatrix},\\
        \mathbf{P}_j^{(\mathrm{row})}
        &=
        \begin{pmatrix}
            b_1 & \cdots & b_J\\
            b_j & \cdots & b_j
        \end{pmatrix}.
    \end{aligned}
\end{equation}

\paragraph{Radial spokes.}
We define $J$ candidate spokes with uniformly spaced orientations $\theta_j=\pi(j-1)/J$, where $j=1,\ldots,J$. For each spoke, we sample $L$ equally spaced values $u_1,\ldots,u_L$ in $[-\kappa,\kappa]$, including both endpoints, with $\kappa=\pi(1-1/J)$. Its coordinate matrix is
\begin{equation}
    \mathbf{P}_j^{(\mathrm{rad})}
    =
    \frac{
        \begin{pmatrix}
            \cos\theta_j\\
            \sin\theta_j
        \end{pmatrix}
        \begin{pmatrix}
            u_1 & \cdots & u_L
        \end{pmatrix}
    }{\max(|\cos\theta_j|,|\sin\theta_j|)}.
\end{equation}
The orientation-dependent scaling extends each spoke to the boundary of the square $[-\kappa,\kappa]^2$.

\subsection{Fourier and Adjoint Operators}
\label{app:operators}
We evaluate the Fourier transform at the selected coordinates using a single-coil nonuniform fast Fourier transform (NUFFT) and implement $\mathcal{G}_{\mathbf{P}_t}$ as its density-compensated adjoint:
\begin{equation}
    \begin{aligned}
        \mathbf{q}_t
        &=\mathcal{F}_{\mathbf{P}_t}\mathbf{x},\\
        \hat{\mathbf{x}}_t
        &=\mathcal{G}_{\mathbf{P}_t}(\mathbf{q}_t)
        =\mathcal{F}_{\mathbf{P}_t}^{*}\mathbf{D}_t\mathbf{q}_t,
    \end{aligned}
\end{equation}
where $\mathbf{D}_t\in\mathbb{R}^{tL\times tL}$ is a diagonal matrix of density-compensation weights that account for nonuniform $k$-space sampling density in radial and mixed-geometry trajectories. We implement both operators with \texttt{TorchKbNufft}~\cite{muckley2020torchkbnufft} using orthonormal scaling, an oversampled grid of size $256\times256$, six interpolation neighbors per dimension, and one iteration of its iterative Pipe--Menon density-compensation estimator~\cite{pipe1999sampling}.

%====================================
% Data and Preprocessing
%====================================

\section{Data and Preprocessing}
\label{app:data}

\subsection{Label Definitions and Data Splits}
\label{app:datasets}
We provide label-assignment and data-split details that complement the main paper. For both datasets, we construct training and validation sets from the official fastMRI~\cite{zbontar2018fastmri} training release and use the official validation release for testing, since fastMRI+~\cite{zhao2022fastmri+} does not provide pathology annotations for the official test sets. Table~\ref{tab:cohorts} reports cohort composition and class prevalences.

\paragraph{Knee tasks.}
For \emph{Meniscus Tear} and \emph{Joint Effusion}, we label a slice as positive if it carries the corresponding image-level annotation and negative otherwise. The combined knee task assigns a positive label when either finding is present.

\paragraph{Brain task.}
Following Schlimbach et al.~\cite{schlimbach2026generative}, the positive class includes seven image-level findings: \emph{mass}, \emph{extra-axial mass}, \emph{nonspecific lesion}, \emph{nonspecific white matter lesion}, \emph{encephalomalacia}, \emph{lacunar infarct}, and \emph{likely cysts}. A slice is positive if any of these findings is annotated, irrespective of additional annotations. Otherwise, a slice is negative if its only image-level annotation is \emph{normal variant}, or if it has no image-level annotations and belongs to a study labeled \emph{normal for age}. We exclude all remaining slices.

%===================================================
% Cohorts
%===================================================

\begin{table}[t]
    \centering
    \caption{\textbf{Cohort composition.} Volume and slice counts by split, with positive slice counts and prevalences (\%) for each task. Brain volume counts include volumes for which all slices are excluded by the label-filtering criteria.}
    \label{tab:cohorts}
    \small
    \setlength{\tabcolsep}{2pt}
    \renewcommand{\arraystretch}{1.12}
    \begin{tabular*}{\columnwidth}{@{\extracolsep{\fill}}lrrr@{}}
        \toprule
        & \textbf{Train} & \textbf{Validation} & \textbf{Test} \\
        \midrule
        \multicolumn{4}{c}{\textbf{Knee}} \\
        \addlinespace[2pt]
        Volumes        & 856              & 117             & 199             \\
        Slices         & 30,562           & 4,180           & 7,135           \\
        \cmidrule(lr){1-4}
        Meniscus tear  & 3,657 (11.97\%)  & 501 (11.99\%)   & 700 (9.81\%)    \\
        Joint effusion & 736 (2.41\%)     & 96 (2.30\%)     & 171 (2.40\%)    \\
        Combined       & 4,354 (14.25\%)  & 584 (13.97\%)   & 864 (12.11\%)   \\
        \midrule
        \multicolumn{4}{c}{\textbf{Brain}} \\
        \addlinespace[2pt]
        Volumes        & 603              & 151             & 247             \\
        Slices         & 4,677            & 975             & 1,861           \\
        \cmidrule(lr){1-4}
        Any abnormality & 1,011 (21.62\%) & 208 (21.33\%)   & 495 (26.60\%)   \\
        \bottomrule
    \end{tabular*}
\end{table}

\subsection{MRI Preprocessing and Augmentation}
\label{app:preprocessing}
For experiments in our retrospective acquisition environment, we use root-sum-of-squares (RSS) magnitude reconstructions, center-pad slices to a square when necessary, and resize them to $128\times128$ using bilinear interpolation. During all training runs, we use TorchIO~\cite{perezgarcia2021torchio} to apply random affine transformations with probability 0.75, using rotations within $\pm7.5^\circ$ and translations within $\pm8$ pixels along each axis. We also apply vertical flipping with probability 0.5. We assign zero phase to the processed images and synthesize measurements using the single-coil Fourier model.

%====================================
% Network Architectures and Training
%====================================

\section{Network Architectures and Training}
\label{app:networks}

\subsection{Diagnostic Classifier}
\label{app:classifier-training}
We use an ImageNet-pretrained ResNet-50~\cite{he2016deep} with magnitude and phase inputs and a linear binary classification head. We adapt each filter in the first convolution by averaging its pretrained RGB kernels and duplicating the resulting kernel for the two inputs. For each image, we independently clip magnitude and phase to their respective 0.5th--99.5th percentile ranges and apply per-channel z-score normalization.

During classifier training, we uniformly sample an acceleration factor $R\in\{1,4,8,16,32\}$ for each batch, setting the readout budget to $T=128/R$. For each sample, we independently choose between a mixed-geometry pattern with probability $\rho_{\mathrm{mix}}=0.5$ and a single uniformly selected geometry otherwise. For mixed patterns, we draw the proportions of Cartesian columns, Cartesian rows, and radial spokes from $\operatorname{Dirichlet}(0.5,0.5,0.5)$ and use largest-remainder rounding to allocate exactly $T$ readouts across geometries.

Within each geometry, we sample readouts without replacement. For Cartesian columns and rows, we choose with equal probability between uniform sampling and a center-weighted variable-density prior with polynomial decay of degree $4$. For radial patterns, we sample spokes uniformly and shift them by a shared random angular offset from $\mathcal{U}(0,\pi/J)$.

We minimize binary focal loss~\cite{lin2017focal} with $\gamma=2$ and construct training batches with a target positive proportion of 25\%. We use AdamW with a learning rate of $5\times10^{-3}$, weight decay of $10^{-4}$, batch size of 1,024, and gradient-norm clipping at 1.0. We linearly increase the learning rate from $5\times10^{-4}$ to $5\times10^{-3}$ over ten warmup epochs, then apply cosine decay to $10^{-4}$. We train for at most 512 epochs, stop after 50 epochs without validation improvement, and select the checkpoint maximizing mean validation AUPRC across all three readout families and $R\in\{4,8,16,32\}$.

\subsection{Modality-Specific Tokenizers}
\label{app:tokenizers}
We detail how the modality-specific tokenizers introduced in the main paper map classifier features and acquired $k$-space measurements to 128-dimensional tokens.

\paragraph{Classifier features.}
We extract the diagnostic classifier's feature map before global average pooling, rearrange it from $2048\times4\times4$ to $16\times2048$, and linearly project each spatial feature vector to a 128-dimensional token. We add learned row and column embeddings to encode spatial position.

\paragraph{Acquired $k$-space.}
We encode each acquired readout as one token. Each of its $L$ measurements is represented by seven features: normalized coordinates $(k_x/\pi,k_y/\pi)$, real and imaginary components, magnitude, phase divided by $\pi$, and normalized sample index $\ell/(L-1)$, where $\ell=0,\ldots,L-1$. We apply layer normalization followed by a shared $7\rightarrow128\rightarrow128$ pointwise MLP with GELU activations and dropout 0.1 between its linear layers. We then concatenate the mean and maximum of the measurement embeddings across the readout, yielding a 256-dimensional vector. Layer normalization followed by a linear projection and GELU maps this vector to a 128-dimensional token, producing $t$ tokens after $t$ readouts.

\subsection{Policy Architecture}
\label{app:policy-architecture}

\paragraph{Backbone.}
We adapt the multimodal diffusion transformer (MMDiT)~\cite{esser2024scaling} to process classifier-feature and acquired $k$-space tokens as two separate streams. The backbone comprises two blocks with token dimension 128 and four attention heads. Each block combines joint attention across streams with a separate feed-forward network for each modality, using hidden dimension 512 and GELU activation. We apply dropout 0.1 to attention weights, attention outputs, and feed-forward layers. For acquisition-step conditioning, we encode the step as a 128-dimensional sinusoidal embedding and project it through a two-layer MLP with width 128 and a SiLU activation. The resulting vector modulates both blocks through adaptive layer normalization. We then apply layer normalization and mean pooling to each output stream and concatenate the two vectors into a 256-dimensional representation.

\paragraph{Readout-scoring head.}
The scoring head applies layer normalization, a $256\rightarrow256$ linear layer, GELU, dropout 0.1, and a final linear projection to one score per candidate readout.

\paragraph{Input ablations.}
For single-modality policies, we omit the unused tokenizer and restrict attention to the active stream. We replace the inactive stream's pooled vector with zeros, preserving the 256-dimensional input to the scoring head. For $k$-space-only policies, a learned token represents the empty initial history. For step-only policies (without patient-dependent inputs), we omit both tokenizers and pass a single constant zero token through the $k$-space stream, retaining acquisition-step conditioning.

\subsection{Algorithm-Specific Networks}
\label{app:algorithm-networks}
We detail the additional network components used by PPO~\cite{schulman2017proximal} and Rainbow~\cite{hessel2018rainbow} in the policy-optimization experiment.
\paragraph{PPO value network.}
We use a separate critic to estimate returns for PPO, retaining the policy's transformer backbone architecture. We concatenate its 256-dimensional pooled representation with separate 32-dimensional embeddings of the normalized acquisition budget and ground-truth label, then pass the resulting 320-dimensional vector through layer normalization, a $320\rightarrow256$ linear layer with GELU activation (without dropout), and a final linear projection to a scalar value. Only the critic receives these additional inputs during training.

\paragraph{Rainbow Q-network.}
For Rainbow, we retain the transformer backbone architecture but disable dropout throughout the network, and replace the scoring head with a dueling distributional head of hidden dimension 256 that predicts a return distribution for each action. Following the original configuration, we use 51 atoms spanning $[-10,10]$ and factorized noisy linear layers with initial noise scale 0.5.

%====================================
% Policy Optimization and Training
%====================================

\section{Policy Optimization Details}
\label{app:optimization}
We describe the implementation and training settings for the RL algorithms evaluated in the policy-optimization experiment. All algorithms use AdamW with learning rate $10^{-4}$, weight decay $10^{-4}$, and $(\beta_1,\beta_2)=(0.9,0.999)$. We train for at most 300 epochs, with early stopping after 50 epochs without validation improvement. We select checkpoints using mean validation AUPRC across the four fixed acceleration factors, $R\in\{4,8,16,32\}$, evaluated on 2,048 validation rollouts for the knee tasks and 975 for the brain task. We collect training rollouts in parallel, with a target positive proportion of 50\%. At inference, we select the highest-scoring readout among those not yet acquired.

\subsection{Greedy}
We follow the multi-sample greedy policy-gradient algorithm used in prior work~\cite{bakker2020experimental,du2024mri}, adopting the hyperparameter notation of Du et al.~\cite{du2024mri}. We collect 64 rollouts in parallel. At each acquisition step, we sample $q=8$ candidate actions with replacement and evaluate their immediate rewards through one-step environment lookahead. We optimize the policy using the resulting multi-sample gradient estimator with minibatches of 512 transitions. We clip gradient norms to 1.0 and reduce the learning rate by a factor of 0.1 every 40 epochs.

\subsection{PPO}
We adapt the CleanRL~\cite{huang2022cleanrl} implementation of PPO~\cite{schulman2017proximal}, retaining its default discount $\gamma=0.99$, GAE parameter $\lambda=0.95$, clipping coefficient 0.2, advantage normalization, clipped value loss with coefficient 0.5, and entropy coefficient 0.01. At each collection iteration, we run 16 rollouts in parallel with a shared acquisition budget sampled uniformly from $\{5,\ldots,32\}$, where 32 corresponds to $R=4$. We perform two passes over all collected transitions, shuffling them before each pass and dividing them into minibatches of up to 256 transitions. We clip gradient norms to 1.0 and reduce the learning rate by a factor of 0.1 every 40 epochs. We detail the value-network architecture in Section~\ref{app:algorithm-networks}.

\subsection{Rainbow}
We adapt the CleanRL~\cite{huang2022cleanrl} implementation of Rainbow~\cite{hessel2018rainbow}, retaining its default three-step returns, discount $\gamma=0.99$, prioritization exponent 0.5, importance-sampling exponent increasing linearly from 0.4 to 1.0, and replay-priority floor $10^{-6}$. At each collection iteration, we run 16 rollouts in parallel with a shared acquisition budget sampled uniformly from $\{5,\ldots,32\}$. We store transitions in a replay buffer with capacity 100,000 and begin optimization after collecting 1,024 transitions. We perform four optimizer steps per collection iteration, each using a replay minibatch of 32 transitions, and update the target network every 1,000 optimizer steps. We use a constant learning rate, clip gradient norms to 10.0, and use the network described in Section~\ref{app:algorithm-networks} with noisy exploration.

%====================================
% Per-Step Probability Calibration
%====================================

\section{Per-Step Probability Calibration}
\label{app:calibration}
Because predicted probabilities determine when acquisition stops, we calibrate them to better reflect observed outcome frequencies. We fit a separate calibration map at each acquisition step to account for changes in classifier calibration as more measurements become available. For each task, method, and seed, we fit these maps using validation predictions only. Given the uncalibrated logit $z_{i,t}$ for slice $i$ at step $t$, we compute
\begin{equation}
    \tilde{p}_{i,t}
    =\sigma\!\left(e^{s_t}z_{i,t}+b_t\right),
\end{equation}
where $\sigma$ denotes the sigmoid function and $s_t$ and $b_t$ are fitted calibration parameters. We minimize mean binary cross-entropy with regularization $10^{-4}(s_t^2+b_t^2)$, initializing $s_t=b_t=0$. We use L-BFGS with learning rate 0.25 and at most 30 iterations, then apply the fitted maps unchanged to test predictions. The positive scale preserves prediction rankings within each acquisition step.

%====================================
% Baseline Implementations
%====================================

\section{Baseline Implementations}
\label{app:baselines}

\subsection{LOUPE-ST and DPS}
\label{app:shared-sampling-baselines}
We adapt LOUPE-ST~\cite{zhang2020extending} and DPS~\cite{huijben2020deep} to pathology prediction by replacing their reconstruction networks with a diagnostic classifier. We jointly optimize the classifier and a shared Cartesian column sampling pattern, propagating classification-loss gradients to the sampling parameters through each method's differentiable sampling mechanism. We retain the sampling formulations, gradient estimators, and hyperparameters described in the respective papers. We use the classifier architecture and optimization settings in Section~\ref{app:classifier-training}, except for an initial learning rate of $10^{-3}$. During training, each method generates Cartesian undersampling masks from its current sampling parameters to construct the classifier inputs. We train a separate sampling-pattern--classifier pair for each task, seed, and acceleration factor. After training, we obtain a fixed sampling pattern by selecting the highest-scoring columns to match the prescribed readout budget. We report fixed-budget results only, as each pattern is optimized for a specific budget.

\subsection{Policy Classifier and ASMR}
\label{app:adaptive-baselines}
We integrate PC~\cite{du2024mri} and ASMR~\cite{yen2024adaptive} into our acquisition environment, following their original implementations unless otherwise specified. Both methods select Cartesian columns only, so we use FFT and inverse FFT operators rather than the NUFFT-based operators in Section~\ref{app:operators}. PC uses the multi-sample greedy policy-gradient algorithm~\cite{bakker2020experimental} described in Section~\ref{app:optimization}, with $q=8$. Following Du et al.~\cite{du2024mri}, we train its ResNet-50 diagnostic classifier using only uniformly sampled Cartesian columns and condition readout selection exclusively on frozen classifier features, which feed a convolutional policy. ASMR optimizes readout selection using PPO~\cite{schulman2017proximal}, retaining the hyperparameters and reward formulation from the authors' implementation. It uses KSpaceNet~\cite{singhal2023feasibility} to encode acquired $k$-space measurements for policy learning and pathology prediction, with its diagnostic classifier trained on variable-density Cartesian columns. For consistency with the other methods, we use a ResNet-50 diagnostic backbone for KSpaceNet. We refer readers to the original papers and implementations of PC\footnote{\url{https://github.com/vios-s/MRI_Active_Sampling}} and ASMR\footnote{\label{fn:asmr-code}\url{https://github.com/robinyen/asmr}} for further details.

\subsection{Fully Sampled Reference}
\label{app:fully-sampled}
We train a separate diagnostic classifier (Section~\ref{app:classifier-training}) on fully sampled images to provide a diagnostic-performance reference on images without undersampling artifacts. Table~\ref{tab:oracle} reports its test performance across five seeds.

\begin{table}[ht]
    \centering
    \small
    \caption{\textbf{Fully sampled reference.} Test AUROC and AUPRC (\%), reported as mean $\pm$ SD across five seeds.}
    \label{tab:oracle}
    \setlength{\tabcolsep}{4pt}
    \renewcommand{\arraystretch}{1.1}
    \begin{tabular*}{\columnwidth}{@{\extracolsep{\fill}}lcc@{}}
        \toprule
        Task & AUROC & AUPRC \\
        \midrule
        Meniscus tear & $92.71 \pm 1.45$ & $60.95 \pm 4.67$ \\
        Joint effusion & $92.33 \pm 2.06$ & $57.25 \pm 4.02$ \\
        Brain abnormality & $84.32 \pm 1.62$ & $73.25 \pm 3.71$ \\
        \bottomrule
    \end{tabular*}
\end{table}

\section{ASMR Experiment Implementation Details}
\label{app:asmr-ablation}
We provide additional details on the experimental setup and the budget-only policy for the ASMR comparison in \cref{sec:asmr-conditioning-results} of the main paper. We use ASMR's official implementation\textsuperscript{\ref{fn:asmr-code}} with its default knee MRI settings and our patient-level knee split. For the budget-only variant, we remove the measurement encoders from the actor and critic and feed a one-hot encoding of the remaining readout budget to separate MLPs with 256-unit hidden layers. Following ASMR~\cite{yen2024adaptive}, we evaluate both policies at sampling rates of 5\%, 8\%, 10\%, and 12.5\%, corresponding to 20, 32, 40, and 50 Cartesian columns, respectively, and select checkpoints by mean validation AUROC.

\section{Additional Results}
\label{app:add-results}
\begin{figure*}[t]
    \centering
    \includegraphics[width=\linewidth]{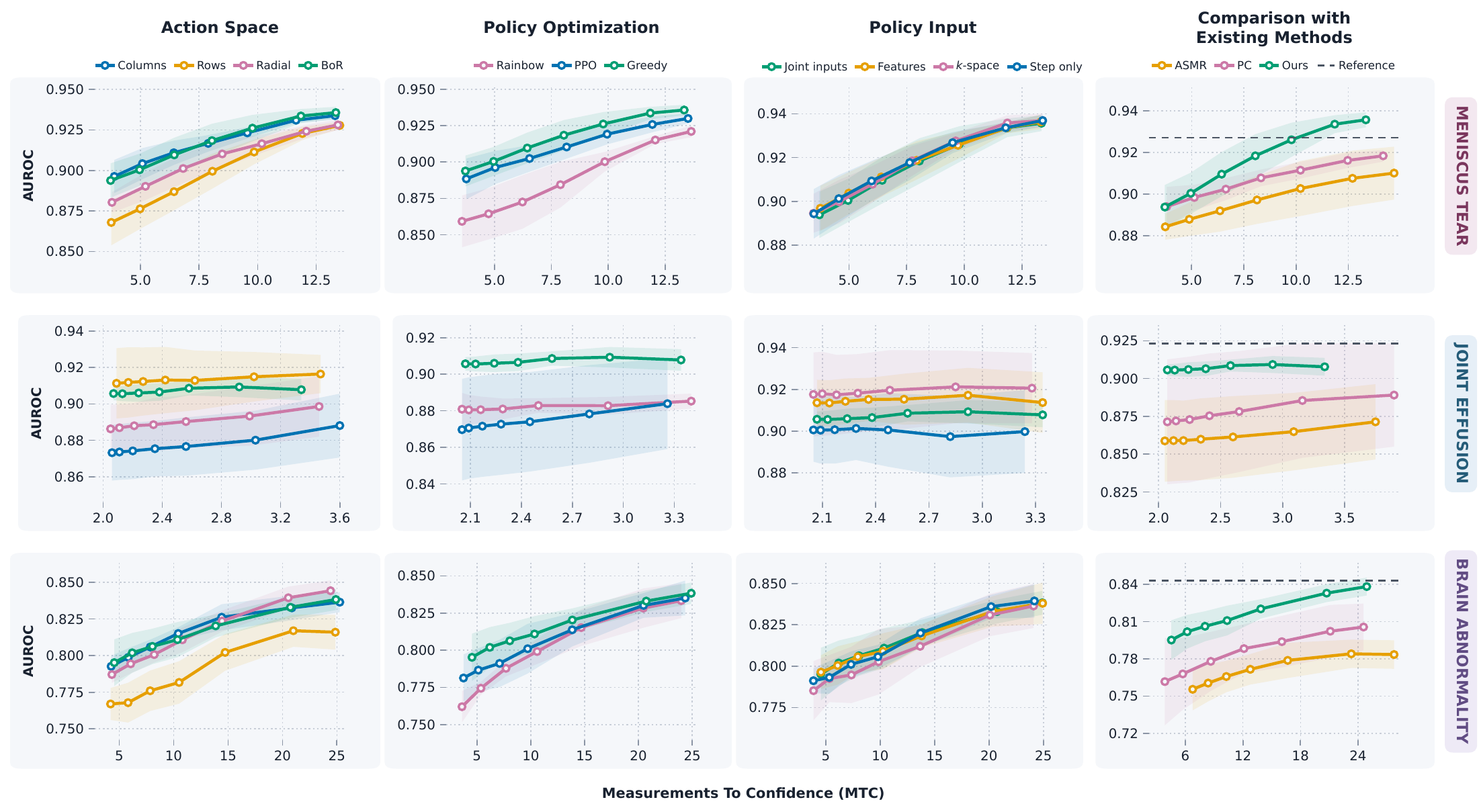}
    \caption{\textbf{AUROC under confidence-based stopping.} AUROC (mean $\pm$ SD) versus MTC for the action-space, policy-optimization, policy-input, and existing-method comparisons (columns). Rows show meniscus-tear, joint-effusion, and brain-abnormality detection. Dashed lines in the final column indicate the fully sampled reference.}
    \label{fig:auroc-mtc-comparisons}
\end{figure*}

\begin{figure*}
    \centering
    \includegraphics[width=0.5\linewidth]{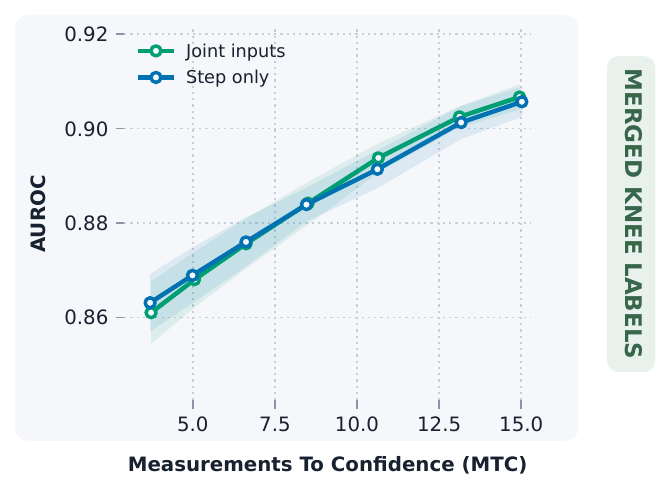}
    \caption{\textbf{Confidence-based stopping on the combined knee task.} AUROC (mean $\pm$ SD) versus MTC for joint-input and step-only policies detecting meniscus tear or joint effusion.}
    \label{fig:merged-knee-auroc-mtc}
\end{figure*}

To complement the graphical results in the main manuscript (\cref{fig:readout-geometry,fig:rl,fig:conditioning,fig:sota}), we report the complete numerical results for the fixed-acceleration evaluations ($R \in \{32, 16, 8, 4\}$) in \cref{tab:fixed-action-space,tab:fixed-optimization,tab:fixed-inputs,tab:fixed-sota}. We also provide the corresponding confidence-based stopping curves ($\eta \in [0.70, 0.975]$) evaluated via AUROC in \cref{fig:auroc-mtc-comparisons,fig:merged-knee-auroc-mtc}. While our primary analysis relies on AUPRC to account for class imbalance, we include AUROC across all tables and curves for completeness, noting that it is less informative when positive-class prevalence is low.

%===================================================
% Fixed acceleration | Action Space
%===================================================

\begin{table*}
    \centering
    \footnotesize
    \setlength{\tabcolsep}{3pt}
    \renewcommand{\arraystretch}{1.12}
    \caption{\textbf{Action-space comparison.} Test AUROC and AUPRC (\%), reported as mean $\pm$ sample standard deviation across five seeds. Acceleration factors $R=32,16,8,4$ correspond to $4,8,16,32$ readouts.}
    \label{tab:fixed-action-space}
    \begin{tabular*}{\textwidth}{@{\extracolsep{\fill}}l*{8}{c}@{}}
        \toprule
        & \multicolumn{2}{c}{$R=32$}
        & \multicolumn{2}{c}{$R=16$}
        & \multicolumn{2}{c}{$R=8$}
        & \multicolumn{2}{c}{$R=4$} \\
        \cmidrule(lr){2-3}
        \cmidrule(lr){4-5}
        \cmidrule(lr){6-7}
        \cmidrule(lr){8-9}
        Action space & AUROC & AUPRC & AUROC & AUPRC & AUROC & AUPRC & AUROC & AUPRC \\
        \midrule
        \multicolumn{9}{c}{Meniscus tear} \\
        \addlinespace[2pt]
        Columns
        & $90.97 \pm 0.77$ & $47.85 \pm 3.61$
        & $92.18 \pm 0.38$ & $52.63 \pm 1.56$
        & $93.02 \pm 0.28$ & $57.49 \pm 0.70$
        & $93.46 \pm 0.32$ & $59.78 \pm 1.70$ \\
        Rows
        & $88.68 \pm 0.62$ & $38.80 \pm 1.65$
        & $91.19 \pm 0.18$ & $46.88 \pm 2.85$
        & $92.20 \pm 0.37$ & $53.22 \pm 4.19$
        & $93.34 \pm 0.20$ & $58.74 \pm 2.35$ \\
        Radial
        & $89.46 \pm 0.75$ & $40.84 \pm 3.26$
        & $90.97 \pm 0.60$ & $44.84 \pm 2.86$
        & $92.41 \pm 0.41$ & $51.94 \pm 3.39$
        & $93.01 \pm 0.37$ & $56.72 \pm 2.66$ \\
        BoR
        & $90.87 \pm 0.59$ & $45.44 \pm 2.51$
        & $92.29 \pm 0.37$ & $51.90 \pm 2.47$
        & $93.13 \pm 0.47$ & $57.51 \pm 3.27$
        & $93.73 \pm 0.32$ & $61.44 \pm 2.08$ \\
        \midrule
        \multicolumn{9}{c}{Joint effusion} \\
        \addlinespace[2pt]
        Columns
        & $91.19 \pm 1.47$ & $48.71 \pm 3.51$
        & $93.37 \pm 1.06$ & $52.86 \pm 2.79$
        & $94.34 \pm 1.14$ & $55.70 \pm 2.79$
        & $94.60 \pm 1.52$ & $56.03 \pm 1.72$ \\
        Rows
        & $91.81 \pm 1.01$ & $52.82 \pm 1.13$
        & $93.13 \pm 0.93$ & $56.62 \pm 3.00$
        & $94.37 \pm 1.49$ & $58.46 \pm 2.73$
        & $95.32 \pm 2.19$ & $59.10 \pm 1.47$ \\
        Radial
        & $91.89 \pm 1.08$ & $48.33 \pm 7.73$
        & $93.77 \pm 0.89$ & $55.52 \pm 3.25$
        & $94.98 \pm 0.52$ & $58.46 \pm 1.57$
        & $95.50 \pm 0.73$ & $59.01 \pm 1.43$ \\
        BoR
        & $91.79 \pm 1.55$ & $53.47 \pm 3.32$
        & $93.33 \pm 2.32$ & $57.20 \pm 1.96$
        & $94.36 \pm 2.56$ & $58.41 \pm 2.25$
        & $94.66 \pm 2.80$ & $58.33 \pm 1.56$ \\
        \midrule
        \multicolumn{9}{c}{Brain abnormality} \\
        \addlinespace[2pt]
        Columns
        & $81.32 \pm 0.69$ & $67.89 \pm 1.35$
        & $82.89 \pm 1.18$ & $71.11 \pm 2.10$
        & $83.85 \pm 0.89$ & $74.02 \pm 2.05$
        & $83.93 \pm 0.69$ & $73.76 \pm 2.00$ \\
        Rows
        & $78.03 \pm 0.66$ & $61.26 \pm 1.57$
        & $79.13 \pm 0.55$ & $64.40 \pm 1.33$
        & $80.30 \pm 0.49$ & $67.08 \pm 1.02$
        & $81.54 \pm 0.82$ & $69.28 \pm 1.80$ \\
        Radial
        & $80.58 \pm 0.61$ & $65.78 \pm 0.99$
        & $82.39 \pm 0.65$ & $69.99 \pm 1.16$
        & $83.50 \pm 0.75$ & $72.40 \pm 1.42$
        & $84.33 \pm 0.49$ & $73.81 \pm 0.88$ \\
        BoR
        & $80.82 \pm 0.56$ & $66.77 \pm 1.48$
        & $82.30 \pm 0.85$ & $70.56 \pm 1.67$
        & $83.44 \pm 0.82$ & $72.70 \pm 1.85$
        & $83.94 \pm 0.59$ & $73.31 \pm 1.46$ \\
        \bottomrule
    \end{tabular*}
\end{table*}

%===================================================
% Fixed acceleration | Policy Optimization
%===================================================

\begin{table*}
    \centering
    \footnotesize
    \setlength{\tabcolsep}{3pt}
    \renewcommand{\arraystretch}{1.12}
    \caption{\textbf{Policy-optimization comparison.} Test AUROC and AUPRC (\%), reported as mean $\pm$ sample standard deviation across five seeds. Acceleration factors $R=32,16,8,4$ correspond to $4,8,16,32$ readouts.}
    \label{tab:fixed-optimization}
    \begin{tabular*}{\textwidth}{@{\extracolsep{\fill}}l*{8}{c}@{}}
        \toprule
        & \multicolumn{2}{c}{$R=32$}
        & \multicolumn{2}{c}{$R=16$}
        & \multicolumn{2}{c}{$R=8$}
        & \multicolumn{2}{c}{$R=4$} \\
        \cmidrule(lr){2-3}
        \cmidrule(lr){4-5}
        \cmidrule(lr){6-7}
        \cmidrule(lr){8-9}
        Algorithm & AUROC & AUPRC & AUROC & AUPRC & AUROC & AUPRC & AUROC & AUPRC \\
        \midrule
        \multicolumn{9}{c}{Meniscus tear} \\
        \addlinespace[2pt]
        Rainbow
        & $88.47 \pm 1.36$ & $39.52 \pm 3.59$
        & $90.51 \pm 1.14$ & $44.41 \pm 2.35$
        & $91.88 \pm 0.66$ & $51.81 \pm 2.08$
        & $92.80 \pm 0.38$ & $56.37 \pm 1.84$ \\
        PPO
        & $90.63 \pm 1.12$ & $46.78 \pm 3.39$
        & $92.18 \pm 0.89$ & $52.26 \pm 3.86$
        & $92.71 \pm 0.63$ & $55.63 \pm 3.67$
        & $93.24 \pm 0.29$ & $58.92 \pm 1.81$ \\
        Greedy
        & $90.87 \pm 0.59$ & $45.44 \pm 2.51$
        & $92.29 \pm 0.37$ & $51.90 \pm 2.47$
        & $93.13 \pm 0.47$ & $57.51 \pm 3.27$
        & $93.73 \pm 0.32$ & $61.44 \pm 2.08$ \\
        \midrule
        \multicolumn{9}{c}{Joint effusion} \\
        \addlinespace[2pt]
        Rainbow
        & $89.83 \pm 2.54$ & $47.06 \pm 4.33$
        & $93.49 \pm 0.91$ & $55.26 \pm 3.08$
        & $95.38 \pm 0.65$ & $57.57 \pm 1.77$
        & $95.69 \pm 0.74$ & $58.13 \pm 0.94$ \\
        PPO
        & $91.39 \pm 2.62$ & $53.65 \pm 1.42$
        & $93.61 \pm 1.50$ & $58.62 \pm 3.31$
        & $95.11 \pm 1.19$ & $58.26 \pm 3.69$
        & $95.70 \pm 1.08$ & $59.31 \pm 1.99$ \\
        Greedy
        & $91.79 \pm 1.55$ & $53.47 \pm 3.32$
        & $93.33 \pm 2.32$ & $57.20 \pm 1.96$
        & $94.36 \pm 2.56$ & $58.41 \pm 2.25$
        & $94.66 \pm 2.80$ & $58.33 \pm 1.56$ \\
        \midrule
        \multicolumn{9}{c}{Brain abnormality} \\
        \addlinespace[2pt]
        Rainbow
        & $78.84 \pm 1.42$ & $63.03 \pm 3.29$
        & $80.78 \pm 1.87$ & $67.18 \pm 3.09$
        & $82.55 \pm 1.62$ & $70.69 \pm 2.33$
        & $83.26 \pm 1.01$ & $72.18 \pm 1.64$ \\
        PPO
        & $80.16 \pm 0.85$ & $66.32 \pm 1.90$
        & $81.77 \pm 0.93$ & $69.85 \pm 1.87$
        & $82.81 \pm 1.21$ & $71.38 \pm 2.04$
        & $83.26 \pm 0.92$ & $71.96 \pm 1.86$ \\
        Greedy
        & $80.82 \pm 0.56$ & $66.77 \pm 1.48$
        & $82.30 \pm 0.85$ & $70.56 \pm 1.67$
        & $83.44 \pm 0.82$ & $72.70 \pm 1.85$
        & $83.94 \pm 0.59$ & $73.31 \pm 1.46$ \\
        \bottomrule
    \end{tabular*}
\end{table*}

%===================================================
% Fixed acceleration | Policy Input Ablation
%===================================================

\begin{table*}[t]
    \centering
    \footnotesize
    \setlength{\tabcolsep}{3pt}
    \renewcommand{\arraystretch}{1.12}
    \caption{\textbf{Policy-input comparison.} Test AUROC and AUPRC (\%), reported as mean $\pm$ sample standard deviation across five seeds. Acceleration factors $R=32,16,8,4$ correspond to $4,8,16,32$ readouts.}
    \label{tab:fixed-inputs}
    \begin{tabular*}{\textwidth}{@{\extracolsep{\fill}}l*{8}{c}@{}}
        \toprule
        & \multicolumn{2}{c}{$R=32$}
        & \multicolumn{2}{c}{$R=16$}
        & \multicolumn{2}{c}{$R=8$}
        & \multicolumn{2}{c}{$R=4$} \\
        \cmidrule(lr){2-3}
        \cmidrule(lr){4-5}
        \cmidrule(lr){6-7}
        \cmidrule(lr){8-9}
        Policy inputs & AUROC & AUPRC & AUROC & AUPRC & AUROC & AUPRC & AUROC & AUPRC \\
        \midrule
        \multicolumn{9}{c}{Meniscus tear} \\
        \addlinespace[2pt]
        Joint inputs
        & $90.87 \pm 0.59$ & $45.44 \pm 2.51$
        & $92.29 \pm 0.37$ & $51.90 \pm 2.47$
        & $93.13 \pm 0.47$ & $57.51 \pm 3.27$
        & $93.73 \pm 0.32$ & $61.44 \pm 2.08$ \\
        Features only
        & $91.15 \pm 0.66$ & $47.87 \pm 3.05$
        & $92.42 \pm 0.20$ & $53.30 \pm 1.69$
        & $93.21 \pm 0.18$ & $57.19 \pm 1.49$
        & $93.75 \pm 0.27$ & $60.90 \pm 2.10$ \\
        $k$-space only
        & $91.07 \pm 0.55$ & $47.72 \pm 1.57$
        & $92.25 \pm 0.40$ & $52.78 \pm 2.20$
        & $93.25 \pm 0.25$ & $58.47 \pm 1.79$
        & $93.82 \pm 0.30$ & $62.07 \pm 1.69$ \\
        Step only
        & $91.05 \pm 0.61$ & $47.87 \pm 1.71$
        & $92.60 \pm 0.32$ & $54.17 \pm 1.83$
        & $93.34 \pm 0.27$ & $58.64 \pm 1.99$
        & $93.80 \pm 0.31$ & $61.43 \pm 1.94$ \\
        \midrule
        \multicolumn{9}{c}{Joint effusion} \\
        \addlinespace[2pt]
        Joint inputs
        & $91.79 \pm 1.55$ & $53.47 \pm 3.32$
        & $93.33 \pm 2.32$ & $57.20 \pm 1.96$
        & $94.36 \pm 2.56$ & $58.41 \pm 2.25$
        & $94.66 \pm 2.80$ & $58.33 \pm 1.56$ \\
        Features only
        & $92.48 \pm 1.16$ & $54.84 \pm 2.21$
        & $93.25 \pm 1.29$ & $57.23 \pm 2.46$
        & $94.49 \pm 1.60$ & $58.09 \pm 2.06$
        & $95.37 \pm 1.29$ & $57.59 \pm 1.10$ \\
        $k$-space only
        & $92.36 \pm 1.48$ & $52.82 \pm 3.38$
        & $94.09 \pm 1.23$ & $56.99 \pm 3.69$
        & $95.58 \pm 1.00$ & $58.96 \pm 1.62$
        & $96.38 \pm 0.71$ & $59.07 \pm 0.83$ \\
        Step only
        & $92.18 \pm 1.15$ & $54.10 \pm 5.16$
        & $93.30 \pm 1.53$ & $56.58 \pm 3.36$
        & $94.83 \pm 0.85$ & $59.06 \pm 2.00$
        & $95.55 \pm 0.67$ & $58.88 \pm 1.81$ \\
        \midrule
        \multicolumn{9}{c}{Brain abnormality} \\
        \addlinespace[2pt]
        Joint inputs
        & $80.82 \pm 0.56$ & $66.77 \pm 1.48$
        & $82.30 \pm 0.85$ & $70.56 \pm 1.67$
        & $83.44 \pm 0.82$ & $72.70 \pm 1.85$
        & $83.94 \pm 0.59$ & $73.31 \pm 1.46$ \\
        Features only
        & $81.14 \pm 0.51$ & $67.10 \pm 1.41$
        & $82.30 \pm 0.42$ & $70.48 \pm 0.39$
        & $83.28 \pm 0.75$ & $72.20 \pm 1.68$
        & $83.91 \pm 1.08$ & $73.15 \pm 2.40$ \\
        $k$-space only
        & $80.13 \pm 0.88$ & $65.27 \pm 1.21$
        & $81.92 \pm 1.05$ & $69.91 \pm 1.99$
        & $83.19 \pm 0.91$ & $72.32 \pm 1.42$
        & $83.62 \pm 0.93$ & $73.34 \pm 1.57$ \\
        Step only
        & $80.55 \pm 0.70$ & $66.06 \pm 1.59$
        & $82.51 \pm 0.53$ & $71.23 \pm 1.43$
        & $83.39 \pm 0.51$ & $72.81 \pm 1.11$
        & $83.84 \pm 0.77$ & $73.10 \pm 1.26$ \\
        \midrule
        \multicolumn{9}{c}{Merged knee labels} \\
        \addlinespace[2pt]
        Joint inputs
        & $88.58 \pm 0.41$ & $49.57 \pm 2.05$
        & $89.80 \pm 0.25$ & $53.29 \pm 1.15$
        & $90.76 \pm 0.37$ & $57.79 \pm 1.90$
        & $91.36 \pm 0.31$ & $59.98 \pm 1.23$ \\
        Step only
        & $88.27 \pm 0.28$ & $49.19 \pm 0.73$
        & $89.99 \pm 0.12$ & $54.24 \pm 1.23$
        & $90.81 \pm 0.33$ & $58.45 \pm 1.78$
        & $91.35 \pm 0.40$ & $60.01 \pm 2.01$ \\
        \bottomrule
    \end{tabular*}
\end{table*}

%===================================================
% Fixed acceleration | Comparison with SOTA
%===================================================

\begin{table*}[t]
    \centering
    \footnotesize
    \setlength{\tabcolsep}{4pt}
    \renewcommand{\arraystretch}{1.12}
    \caption{\textbf{Comparison with existing methods.} Test AUROC and AUPRC (\%), reported as mean $\pm$ sample standard deviation across five seeds. Acceleration factors $R=32,16,8,4$ correspond to $4,8,16,32$ readouts. Under strong undersampling ($R\in\{32,16\}$), joint optimization in LOUPE-ST and DPS can collapse to chance-level predictions, yielding zero standard deviation when all five seeds collapse at $R=32$. Both methods converge reliably across all five seeds at $R\in\{8,4\}$.}
    \label{tab:fixed-sota}
    \begin{tabular}{l*{8}{c}}
        \toprule
        & \multicolumn{2}{c}{$R=32$}
        & \multicolumn{2}{c}{$R=16$}
        & \multicolumn{2}{c}{$R=8$}
        & \multicolumn{2}{c}{$R=4$} \\
        \cmidrule(lr){2-3}
        \cmidrule(lr){4-5}
        \cmidrule(lr){6-7}
        \cmidrule(lr){8-9}
        Method & AUROC & AUPRC & AUROC & AUPRC & AUROC & AUPRC & AUROC & AUPRC \\
        \midrule
        \multicolumn{9}{c}{Meniscus tear} \\
        \addlinespace[2pt]
        LOUPE-ST~\cite{zhang2020extending}
        & $46.61 \pm 0.00$ & $9.31 \pm 0.00$
        & $57.81 \pm 13.87$ & $13.70 \pm 5.96$
        & $89.85 \pm 0.92$ & $42.54 \pm 3.34$
        & $91.53 \pm 0.77$ & $48.10 \pm 2.27$ \\
        DPS~\cite{huijben2020deep}
        & $50.88 \pm 9.01$ & $10.52 \pm 2.77$
        & $87.64 \pm 2.39$ & $35.72 \pm 5.31$
        & $90.53 \pm 1.16$ & $45.24 \pm 2.55$
        & $91.12 \pm 0.66$ & $46.53 \pm 3.71$ \\
        ASMR~\cite{yen2024adaptive}
        & $89.75 \pm 1.10$ & $42.63 \pm 2.13$
        & $90.55 \pm 1.20$ & $45.22 \pm 2.60$
        & $90.82 \pm 1.26$ & $47.14 \pm 3.87$
        & $91.13 \pm 1.17$ & $48.64 \pm 4.18$ \\
        PC~\cite{du2024mri}
        & $90.20 \pm 0.59$ & $44.99 \pm 2.58$
        & $91.02 \pm 0.35$ & $48.29 \pm 1.39$
        & $91.62 \pm 0.30$ & $50.78 \pm 1.43$
        & $91.94 \pm 0.35$ & $51.54 \pm 1.85$ \\
        Ours
        & $90.87 \pm 0.59$ & $45.44 \pm 2.51$
        & $92.29 \pm 0.37$ & $51.90 \pm 2.47$
        & $93.13 \pm 0.47$ & $57.51 \pm 3.27$
        & $93.73 \pm 0.32$ & $61.44 \pm 2.08$ \\
        \midrule
        \multicolumn{9}{c}{Joint effusion} \\
        \addlinespace[2pt]
        LOUPE-ST~\cite{zhang2020extending}
        & $56.20 \pm 0.00$ & $2.52 \pm 0.00$
        & $72.84 \pm 14.64$ & $12.85 \pm 11.71$
        & $93.40 \pm 1.83$ & $41.58 \pm 5.02$
        & $95.01 \pm 0.57$ & $47.29 \pm 3.23$ \\
        DPS~\cite{huijben2020deep}
        & $65.20 \pm 11.35$ & $5.66 \pm 4.29$
        & $92.92 \pm 0.72$ & $36.46 \pm 3.72$
        & $92.97 \pm 1.28$ & $40.66 \pm 5.17$
        & $94.83 \pm 1.00$ & $48.37 \pm 6.59$ \\
        ASMR~\cite{yen2024adaptive}
        & $89.37 \pm 1.63$ & $43.67 \pm 3.41$
        & $91.10 \pm 1.61$ & $47.39 \pm 3.68$
        & $92.61 \pm 0.70$ & $51.07 \pm 4.83$
        & $93.50 \pm 0.38$ & $53.40 \pm 3.60$ \\
        PC~\cite{du2024mri}
        & $90.69 \pm 1.43$ & $40.71 \pm 9.31$
        & $92.05 \pm 1.06$ & $46.90 \pm 1.64$
        & $93.54 \pm 0.58$ & $50.65 \pm 0.46$
        & $94.38 \pm 0.84$ & $52.54 \pm 2.55$ \\
        Ours
        & $91.79 \pm 1.55$ & $53.47 \pm 3.32$
        & $93.33 \pm 2.32$ & $57.20 \pm 1.96$
        & $94.36 \pm 2.56$ & $58.41 \pm 2.25$
        & $94.66 \pm 2.80$ & $58.33 \pm 1.56$ \\
        \midrule
        \multicolumn{9}{c}{Brain abnormality} \\
        \addlinespace[2pt]
        LOUPE-ST~\cite{zhang2020extending}
        & $48.03 \pm 0.00$ & $28.81 \pm 0.00$
        & $53.53 \pm 12.29$ & $33.71 \pm 10.96$
        & $77.15 \pm 1.39$ & $58.18 \pm 4.36$
        & $78.40 \pm 1.90$ & $61.31 \pm 4.26$ \\
        DPS~\cite{huijben2020deep}
        & $48.03 \pm 0.00$ & $28.81 \pm 0.00$
        & $76.14 \pm 1.96$ & $56.71 \pm 2.96$
        & $75.83 \pm 0.65$ & $55.49 \pm 1.09$
        & $77.25 \pm 2.36$ & $61.53 \pm 4.48$ \\
        ASMR~\cite{yen2024adaptive}
        & $75.80 \pm 2.03$ & $55.53 \pm 3.25$
        & $76.85 \pm 1.53$ & $59.23 \pm 3.19$
        & $78.03 \pm 1.28$ & $61.98 \pm 4.34$
        & $78.35 \pm 1.34$ & $63.15 \pm 3.73$ \\
        PC~\cite{du2024mri}
        & $78.82 \pm 1.87$ & $64.08 \pm 4.19$
        & $79.89 \pm 1.57$ & $66.61 \pm 3.51$
        & $80.67 \pm 1.89$ & $68.73 \pm 3.38$
        & $81.02 \pm 1.95$ & $69.37 \pm 3.29$ \\
        Ours
        & $80.82 \pm 0.56$ & $66.77 \pm 1.48$
        & $82.30 \pm 0.85$ & $70.56 \pm 1.67$
        & $83.44 \pm 0.82$ & $72.70 \pm 1.85$
        & $83.94 \pm 0.59$ & $73.31 \pm 1.46$ \\
        \bottomrule
    \end{tabular}
\end{table*}